%% file: main.tex
\documentclass{article}
\usepackage[T1]{fontenc}
\usepackage{lmodern}
\usepackage{iclr2026_conference,times}

\usepackage{amsmath}
\usepackage{amssymb}
\usepackage{booktabs}
\usepackage{graphicx}
\usepackage{float}
\usepackage{microtype}
\usepackage{multirow}
\usepackage{xcolor}
\usepackage{hyperref}
\usepackage{url}

\usepackage{wrapfig}
\usepackage{tikz}
\usepackage[table]{xcolor}

\usetikzlibrary{positioning,arrows.meta,fit,backgrounds,matrix}

\graphicspath{{figures/}}

\newcommand{\spectra}{\textsc{Spectra}}

\iclrfinalcopy

\title{SPECTRA: Pushing the KV Cache Beyond the 2-Bit Cliff via Spectral Transform Coding}

\author{%
  Jiamu Zhang$^{1,2}$\thanks{Work done while intern at Nokia.}\quad
  Liang Wu$^{1}$\quad
  Kelly Wan$^{1}$\quad
  Hanjie Chen$^{2}$\quad
  Liangjie Hong$^{1}$ \\[3pt]
  $^{1}$Nokia \qquad $^{2}$Rice University%
}

\begin{document}

\maketitle
\lhead{Preprint. Under review.}

\begin{abstract}
\input{sections/00_abstract}
\end{abstract}

\input{sections/01_introduction}

\input{sections/motivation}

\input{sections/03_method}
\input{sections/04_experiments}

\input{sections/02_related}
\input{sections/05_conclusion}

\bibliography{spectra}
\bibliographystyle{iclr2026_conference}

\newpage
\appendix
\input{sections/appendix}

\end{document}

%% file: sections/00_abstract.tex
Large language models (LLMs) increasingly read very long inputs in the agentic era nowadays, from whole documents and codebases to conversations across many turns. Their inference memory is then dominated by the key-value (KV) cache, the running store of the attention keys and values of everything the model has read and generated. Since the cache grows with the context length and is re-read in full at every generated token, a longer context directly means more GPU memory, until the cache dominates what the hardware can hold.

To reduce this cost, most existing methods compress the KV cache by lowering every stored value to the same low precision, a technique known as quantization. They can push this to nearly two bits per value, but rarely further, because at this so-called 2-bit cliff quality drops sharply: four levels per value are too few for the cache's outlier-heavy values, where a few large entries consume the levels and collapse the many small ones into noise. A natural remedy is to spend more bits on the channels that matter and fewer on the rest, but the raw cache offers no handle: its channels are strongly correlated, so none is clearly more important than another.

To fill this gap, we conduct a comprehensive study and find that this handle appears once the cache is rotated into a coordinate system that is computed from its own statistics and removes these correlations. There, the channels' contributions are highly uneven: a small fraction carries almost all of the information, and spending the budget on those few is far more accurate than spreading it evenly.

Guided by these observations, we develop SPECTRA, a training-free, drop-in codec that re-encodes the cache into this coordinate system and concentrates the bit budget on the channels that carry the signal. On Llama-3.1-8B and Qwen2.5-7B over long-context benchmarks, SPECTRA is near-lossless at 4x compression, competitive at 8x where uniform quantization has collapsed, and reaches up to 12x, pushing usable compression past the 2-bit cliff and letting the same GPU serve much longer contexts and larger batches at higher throughput. Code: \url{https://github.com/nokia-applied-research/SPECTRA}.

%% file: sections/01_introduction.tex
\section{Introduction}
\label{sec:intro}

Large language models (LLMs) now power a growing range of agentic applications. A coding assistant may work over an entire repository, and a dialogue agent may run for many turns while calling external tools~\cite{yao2023reactsynergizingreasoningacting, schick2023toolformerlanguagemodelsteach, jimenez2024swebenchlanguagemodelsresolve, yang2024sweagentagentcomputerinterfacesenable}. What these applications have in common is length: the context, together with the model's own generated reasoning, can reach hundreds of thousands of tokens~\cite{geminiteam2024gemini15unlockingmultimodal, hsieh2024rulerwhatsrealcontext, bai2024longbenchbilingualmultitaskbenchmark}. At this scale the main cost of serving is no longer the model weights but the key-value (KV) cache, which stores the attention keys and values of past tokens so the model does not recompute them. Because every new token attends to the whole cache, the cache must sit in fast memory and be reloaded at each step, and its size grows with both the context length and the number of concurrent requests. As a result, at long context the KV cache can grow larger than the model weights themselves~\cite{kwon2023efficientmemorymanagementlarge, zheng2024sglangefficientexecutionstructured}, and it becomes the main constraint on the context length and the number of concurrent requests that the serving hardware can support.

Many methods compress the KV cache, and they reduce it along one of three axes. Token eviction methods reduce the cache by discarding tokens they judge unimportant, such as H2O~\cite{zhang2023h2oheavyhitteroracleefficient} and SnapKV~\cite{li2024snapkvllmknowslooking}, which keep the tokens that have received the most attention, or StreamingLLM~\cite{xiao2024streamingllm}, which keeps the first and the most recent tokens. Low-rank methods reduce the cache by storing keys and values in fewer dimensions, such as Palu~\cite{chang2024palucompressingkvcachelowrank} and Eigen Attention~\cite{saxena2024eigenattentionattentionlowrank}, which project them onto a low-rank subspace obtained from the model's weights or activations. Quantization methods reduce the cache by storing every value at lower precision, such as KIVI~\cite{liu2024kivi} and KVQuant~\cite{hooper2024kvquant}, which quantize keys per channel and values per token to contain outlier channels. These families differ in what they remove, yet most share the same strategy: they fix the budget in advance, as a set of tokens, a target rank, or a bit-width, and rarely adapt it to how much each part of the cache matters for the output. This limitation is clearest for quantization. Lowering every value to the same precision reaches about two bits, but rarely goes further, because at this so-called 2-bit cliff~\cite{dettmers2023case4bitprecisionkbit, zhou2026signaldegradationcomputationcollapse}, accuracy drops sharply: with only four levels per value, a few large entries take up the range and the many small ones collapse into noise~\cite{xiao2024smoothquantaccurateefficientposttraining, dettmers2022llmint88bitmatrixmultiplication}.

We take a different view. Rather than fixing any of these choices in advance, we ask where each bit of the budget should go, and treat KV-cache compression as one problem: how to spend a limited budget on the parts of the cache that matter most. Seen this way, the three axes are settings of a single allocation, and two of them are really one. Low-rank projection removes channels, while quantization sets how many bits each channel keeps, and a removed channel is just a channel kept at zero bits. Choosing a rank and choosing a precision are therefore the same decision: how many bits to give each channel, from several down to zero. To make this decision well, we first need to know which channels matter, yet the raw cache does not reveal this, because its channels are correlated and none stands out on its own. We find that the answer appears once the cache is rotated into its own principal directions, the orthogonal axes along which it varies the most, which we estimate from its own statistics. In this basis the channels are uncorrelated, and a small set of them accounts for most of the cache. 

Building on this, we introduce SPECTRA, a training-free and drop-in codec that rotates the cache into its spectral basis, the principal directions along which it varies, and spends its bit budget on the channels that carry the signal. SPECTRA needs only a small, one-time calibration to estimate this basis, after which its channels are uncorrelated and ordered by how much they contribute. At inference, it stores each key and value in this basis and gives every channel a bit-width that matches its contribution: more bits to the few dominant channels, and fewer, down to zero, for the rest. A channel given zero bits is simply removed, so the single rule that sets each channel's precision also sets its rank. Because the basis is a small fixed matrix that folds into attention, SPECTRA reconstructs the cache with little overhead and without fine-tuning. Our contributions are summarized as follows:
\begin{itemize}
  \item \textbf{An analysis of the KV cache showing that its information is concentrated in a
        few of its principal directions, where a small set of channels carries most of
        the signal.} This recasts KV-cache compression as spending a fixed bit budget on
        the channels that matter, and unifies low-rank projection and quantization as the
        two ends of one control, since a channel given zero bits is simply dropped.
  \item \textbf{SPECTRA, a training-free and drop-in KV-cache codec that transforms the cache
        into these principal directions and allocates its bits to the channels that carry
        the signal.} We evaluate SPECTRA on Llama-3.1-8B, Mistral-7B, and Qwen2.5-7B across
        LongBench and RULER. SPECTRA compresses the KV cache further than existing
        quantization and low-rank methods at the same quality, and stays near-lossless
        well past 8x compression rate at which uniform quantization breaks down.
\end{itemize}

%% file: sections/motivation.tex
\section{Motivation: Where Should Each Bit Go?}
\label{sec:motivation}

We view the KV cache as a signal to be coded rather than a tensor to be trimmed, so as a result, 
compressing it is then a rate--distortion problem, in which the question is not how
many bits to use or which tokens to keep, but where each bit should go. This reframes
KV-cache compression from an engineering heuristic into a coding problem, and raises a
concrete question: given a fixed budget, in which basis and by what measure of
importance should each bit be allocated across channels and precision? We answer it by
probing the KV cache of pretrained LLMs directly, scoring each choice by the error it
induces in the attention output on held-out text; the statistics that define our transform come from a small calibration set. \textbf{The three observations below answer this question in turn.}

\begin{figure}[ht]
  \centering
  \includegraphics[width=0.88\linewidth]{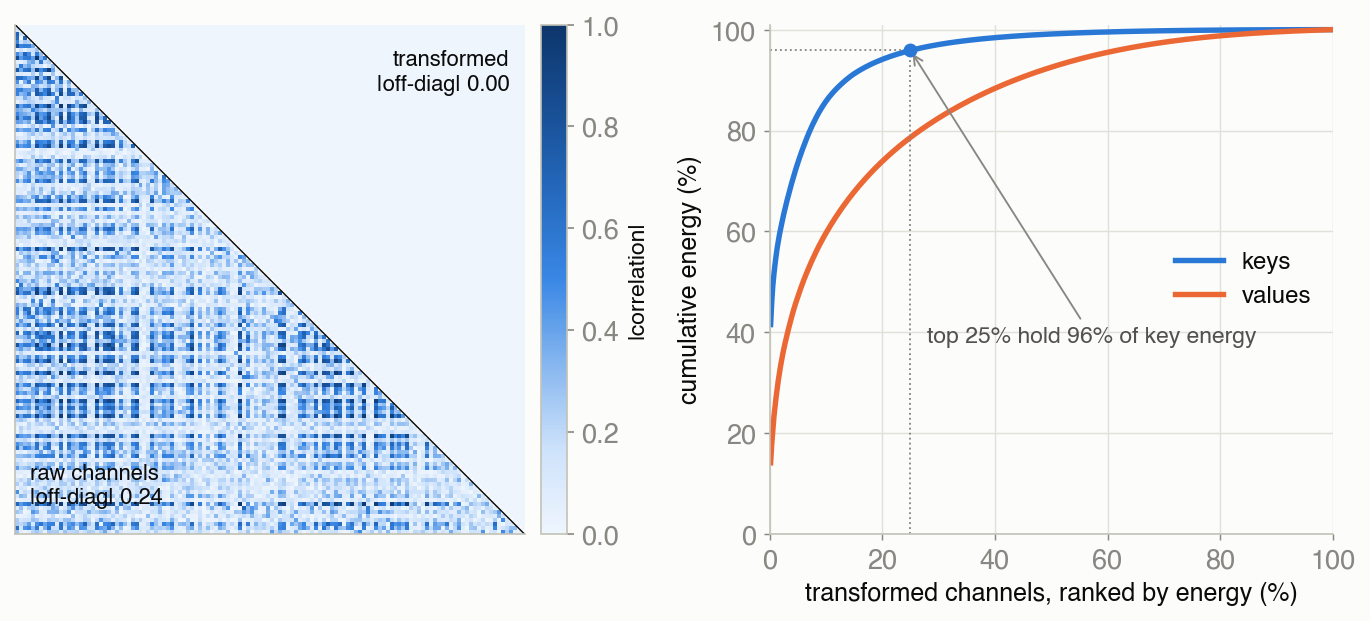}
  \caption{\textbf{Observation 1.}
  \emph{Left:} magnitude of the key channel-correlation matrix; the lower triangle shows the
  raw channels (mean off-diagonal $\approx 0.2$) and the upper triangle shows the channels
  after rotating the cache into a basis computed from its own statistics ($\approx 0$).
  \emph{Right:} cumulative energy over channels ranked by energy; the top $25\%$ of channels
  hold $96\%$ of the key energy, and the value curve is shown for reference.}
  \label{fig:ob1}
\end{figure}

\subsection{\textbf{OB1: The cache's importance is concentrated, but only in the right basis}} 
In the raw channel basis the keys within a layer are strongly correlated, so no channel stands out as more important than another and there is no natural axis to allocate along, lower triangle: mean off-diagonal correlation $\approx 0.2$). Once we
rotate the cache into a basis computed from its own statistics, the channels become
uncorrelated (upper triangle: $\approx 0$), and their energy concentrates sharply: the top
$25\%$ of channels carry $96\%$ of the key energy, and the rest are nearly redundant
(Figure~\ref{fig:ob1}, right). Importance is therefore separable and heavily skewed, but only
after this transform, and this tells us to allocate in the transformed basis rather than the
raw one.

\begin{figure}[ht]
  \centering
  \includegraphics[width=0.88\linewidth]{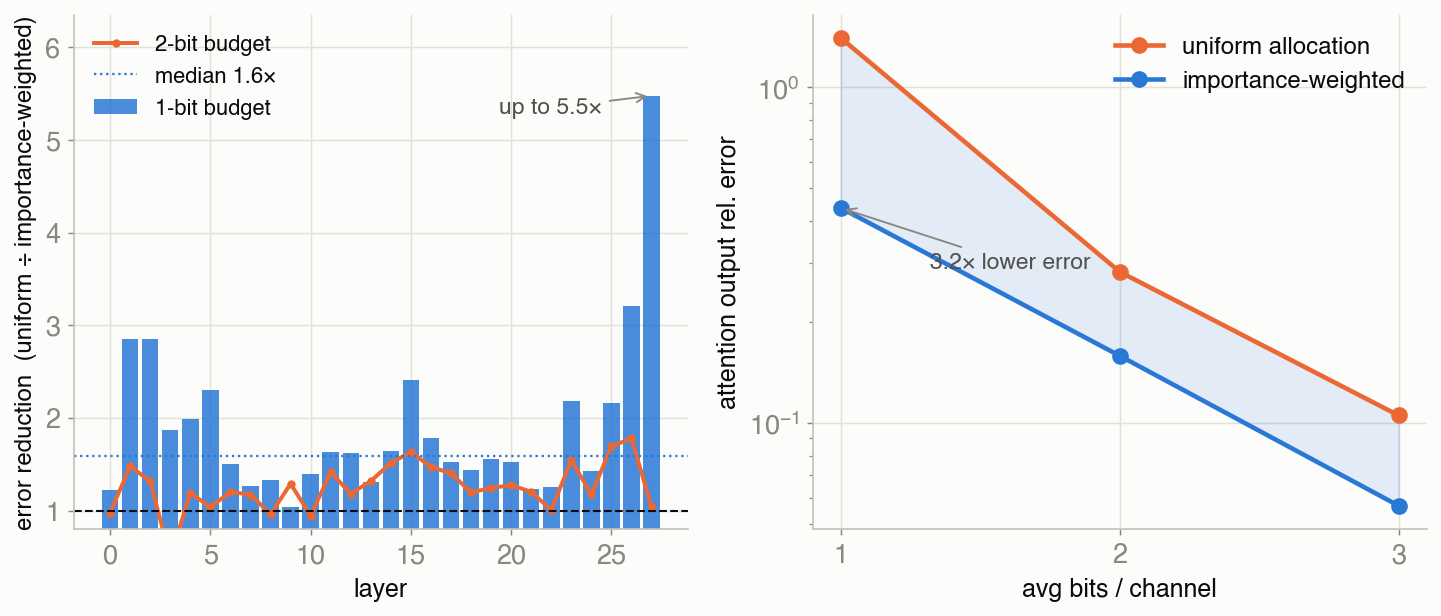}
  \caption{\textbf{Observation 2.} \emph{Left:} per-layer reduction in attention-output error from importance-weighted allocation over uniform (uniform error $\div$ importance-weighted error) at a one-bit and a
  two-bit budget; the reduction exceeds $1\times$ at almost every layer, with a median of
  $1.6\times$ and up to $5.5\times$ at one bit. \emph{Right:} attention-output error versus
  average bits per channel at a representative layer; the gap is largest at one bit and closes
  as precision grows. Channels given zero bits are dropped, so this allocation unifies rank
  reduction and quantization.}
  \label{fig:ob2}
\end{figure}

\subsection{\textbf{OB2: Allocating bits by importance beats uniform precision, most sharply below two bits.}}
Because the channels differ so much in energy, giving each the same number of bits wastes
the budget. When we instead give more bits to high-energy channels and fewer, down to zero,
to low-energy ones, the attention-output error drops at almost every layer (Figure~\ref{fig:ob2},
left), by a median of $1.6\times$ and up to $5.5\times$ at one bit per channel. The
advantage is concentrated where uniform quantization fails: the gap is largest at one bit
and closes as the bit-width grows (Figure~\ref{fig:ob2}, right), exactly the sub-two-bit regime
that gives the 2-bit cliff its name. A channel assigned zero bits is simply dropped, so this
single allocation spans both quantization and rank reduction, making low-rank projection and
quantization the two ends of one control.

\begin{figure}[ht]
  \centering
  \includegraphics[width=0.88\linewidth]{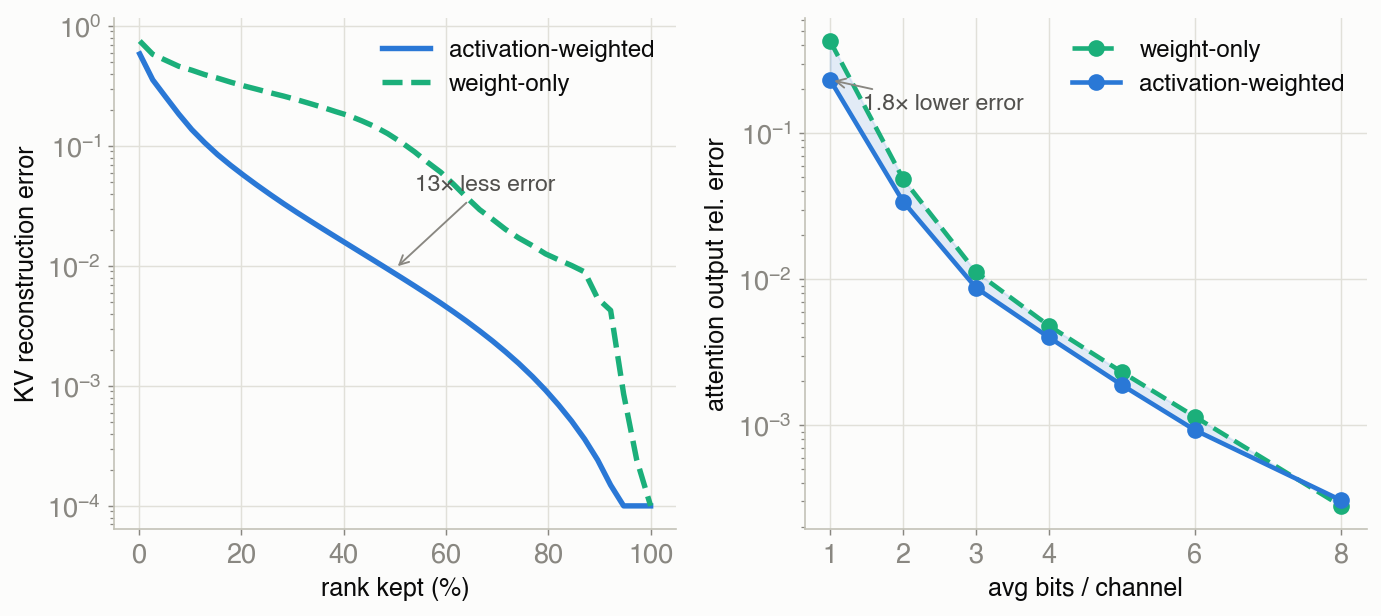}
  \caption{\textbf{Observation 3.} Both panels
  compare an activation-weighted transform (from $\mathbf{G}=\mathbb{E}[\mathbf{h}^{\top}\mathbf{h}]$)
  with a weight-only transform. \emph{Left:} KV reconstruction error versus the fraction of
  rank kept; at half the rank the activation-weighted error is about $13\times$ lower.
  \emph{Right:} attention-output error versus average bits per channel; at a matched budget the
  activation-weighted error is about $1.9\times$ lower.}
  \label{fig:ob3}
\end{figure}

\subsection{\textbf{OB3: A channel's importance is set by the data, not its weights.}}
Wo observe that how much a channel matters depends on how strongly the input activations drive it, not on
the magnitude of its projection weights. When we build the transform and rank the channels
from the activation statistics $\mathbf{G}=\mathbb{E}[\mathbf{h}^{\top}\mathbf{h}]$, where
$\mathbf{h}$ is the layer input to the key/value projection, rather than from the weight
matrix alone, we keep far more of the cache at the same budget: at half the rank the
reconstruction error is about $13\times$ lower (Figure~\ref{fig:ob3}, left), and at a matched
bit-rate the attention-output error is about $1.9\times$ lower (Figure~\ref{fig:ob3}, right). The
basis and the importance measure must therefore both be activation-weighted, not weight-only.

Together these observations specify our method. We encode the cache in an activation-weighted
transform that decorrelates and concentrates it (Observations~1 and~3), and we spend a fixed
bit budget across the resulting channels by importance, letting the weakest channels fall to
zero bits (Observation~2). Section~\ref{sec:method} makes this precise as SPECTRA.

%% file: sections/03_method.tex
\section{Method}
\label{sec:method}

Throughout this section, bold lowercase denotes row vectors (a hidden state $\mathbf h\in\mathbb R^{1\times d}$, a key $\mathbf k\in\mathbb R^{1\times d_{kv}}$, a latent $\mathbf c\in\mathbb R^{1\times r}$) and bold uppercase denotes matrices (the projection $\mathbf W_K\in\mathbb R^{d\times d_{kv}}$, the stacked cache $\mathbf K=\mathbf H\mathbf W_K\in\mathbb R^{L\times d_{kv}}$, and the uncentered second-moment matrix $\mathbf G=\mathbb E_{\mathbf h}[\mathbf h^\top\mathbf h]\in\mathbb R^{d\times d}$); $\mathbb E_{\mathbf h}$ is the expectation over hidden states on calibration data.

Based on the 3 observations mentioned in the previous section, our reframing casts KV-cache compression as deciding where to spend a fixed budget, which leaves one design choice to make concrete: what the cache should store. The answer follows from how redundant the produced keys and values are. Measured in the metric of the model's activations, most of their energy occupies a small subspace, so we cache a compact latent in place of $\mathbf K$ and $\mathbf V$ and recover them only when needed. This single decision organizes the rest of the section. We first build the latent through a $G$-weighted transform (a KLT in the $G$-metric) that is second-moment-orthogonal and energy-ordered by construction (Section~\ref{subsec:latent}); we then quantize it at a rate--distortion optimum that this very structure makes well-posed (Section~\ref{subsec:quant}); and we finally let the same latent turn its memory saving into a compute saving, through attention carried out directly in the latent space (Section~\ref{subsec:system}).

\input{figs/fig_method}

\subsection{Caching the Latent: A \texorpdfstring{$G$}{G}-weighted Transform}
\label{subsec:latent}

We now make the latent precise. We seek a rank-$r$ factorization of each key projection,
$\mathbf W_K \approx \mathbf W_{\text{down}} \mathbf W_{\text{up}}$ with $\mathbf W_{\text{down}}\in\mathbb{R}^{d\times r}$ and
$\mathbf W_{\text{up}}\in\mathbb{R}^{r\times d_{kv}}$ ($r \le d_{kv}$), so that the cache holds the latent
$\mathbf c = \mathbf h \mathbf W_{\text{down}}$ and recovers a key as $\hat{\mathbf k} = \mathbf c\,\mathbf W_{\text{up}}$ only when it is needed.
Over a length-$L$ sequence this replaces the conventional $L\times d_{kv}$ cache $\mathbf K = \mathbf H\mathbf W_K$ with the
$L\times r$ latent $\mathbf C = \mathbf H\mathbf W_{\text{down}}$, reconstructing $\hat{\mathbf K}=\mathbf C\,\mathbf W_{\text{up}}$ on demand
(Figure~\ref{fig:method}(a)); the low-energy directions beyond rank $r$ are simply discarded by the
truncation, so the discarded tail is a rank reduction rather than a separate pruning step.
The question is what ``$\approx$'' should mean here. It should \emph{not} mean that
$\mathbf W_K \approx \mathbf W_{\text{down}}\mathbf W_{\text{up}}$ in the ordinary Frobenius sense, which treats every
direction of the weight as equally important; what we actually care about is that the keys the
model \emph{produces} are preserved, $\mathbf h\mathbf W_K \approx \mathbf h\,\mathbf W_{\text{down}}\mathbf W_{\text{up}}$, on the
activations $\mathbf h$ that actually occur. The right objective is therefore an expectation over the
model's own activation distribution,
\begin{equation}
\min_{\operatorname{rank} r}\ \mathbb{E}_{\mathbf h}\big\| \mathbf h\,(\mathbf W_K - \mathbf W_{\text{down}}\mathbf W_{\text{up}}) \big\|^2 ,
\label{eq:latent-objective}
\end{equation}
and identically for $\mathbf W_V$.

The expectation in Equation~\eqref{eq:latent-objective} has a convenient closed form. Writing
$\Delta\mathbf W = \mathbf W_K - \mathbf W_{\text{down}}\mathbf W_{\text{up}}$ and using $\mathbf G=\mathbb{E}_{\mathbf h}[\mathbf h^\top \mathbf h]$, a short
calculation (Appendix~\ref{app:GSVD}) gives $\mathbb{E}_{\mathbf h}\|\mathbf h\,\Delta\mathbf W\|^2 = \|\mathbf G^{1/2}\Delta\mathbf W\|_F^2$, so
the objective becomes a $G$-weighted low-rank problem,
\begin{equation}
\min_{\operatorname{rank} r}\ \big\| \mathbf G^{1/2}\,(\mathbf W_K - \mathbf W_{\text{down}}\mathbf W_{\text{up}}) \big\|_F .
\label{eq:gweighted}
\end{equation}
The factor $\mathbf G^{1/2}$ simply re-weights the reconstruction error by how often each input direction
is actually used: a direction the model visits constantly is expensive to get wrong, while one it
never visits costs nothing, regardless of how large the corresponding weights are. This is the one
place \spectra{} departs from plain low-rank factorization of the cache, such as Palu~\citep{chang2024palucompressingkvcachelowrank}
or SVD-LLM~\citep{wang2024svdllm}: we minimize error in the metric of the activations, not in raw
weight space. This is exactly the distinction our experiments confirm at the level of end-task accuracy.

Problem~\eqref{eq:gweighted} has a closed-form solution. Whitening the weight (rescaling it by
$\mathbf G^{1/2}$ so that reconstruction error is measured in the activation metric) as $\mathbf M = \mathbf G^{1/2} \mathbf W_K$
and taking its SVD $\mathbf M = \mathbf U\mathbf\Sigma \mathbf V^\top$, the optimal rank-$r$ factor is obtained by keeping the top
$r$ components (Eckart--Young~\citep{Eckart_Young_1936}, the classical result that a truncated SVD is
the best low-rank approximation, now in the $G$-metric) and undoing the whitening with a
\emph{square-root} split of the singular values:
\begin{equation}
\mathbf W_{\text{down}} = \mathbf G^{-1/2}\,\mathbf U_r\,\mathbf\Sigma_r^{1/2},
\qquad
\mathbf W_{\text{up}} = \mathbf\Sigma_r^{1/2}\,\mathbf V_r^\top .
\label{eq:gsvd}
\end{equation}
Any split of $\mathbf\Sigma_r$ between the two factors gives the same reconstruction $\mathbf W_{\text{down}}\mathbf W_{\text{up}}$;
what the square-root split buys is the statistics of the cached quantity itself. Substituting
$\mathbf c = \mathbf h \mathbf W_{\text{down}}$ and using $\mathbb{E}_{\mathbf h}[\mathbf h^\top \mathbf h]=\mathbf G$,
\begin{equation}
\mathbb{E}[\mathbf c^\top \mathbf c]
= \mathbf W_{\text{down}}^\top\,\mathbb{E}[\mathbf h^\top \mathbf h]\,\mathbf W_{\text{down}}
= \mathbf W_{\text{down}}^\top \mathbf G \mathbf W_{\text{down}}
= \mathbf\Sigma_r
\qquad (\text{diagonal, descending}).
\label{eq:second-moment-latent}
\end{equation}
Thus, the latent coordinates are \textbf{orthogonal in uncentered second
moment} and \textbf{energy-ordered}, with coordinate $j$ carrying second
moment $(\mathbf\Sigma_r)_{jj}$. Equation~\eqref{eq:second-moment-latent} should not
be read as a covariance identity when $\mathbb{E}[\mathbf c]\neq 0$: before
quantization, we subtract the measured coordinate means and estimate the
centered marginal variances separately. The resulting canonical ordering
distinguishes the transform from a plain SVD, which does not use the activation
metric, and from the random rotations used by TurboQuant~\cite{zandieh2025turboquantonlinevectorquantization} or QuaRot~\cite{ashkboos2024quarotoutlierfree4bitinference} , which impose
no energy ordering.\footnote{A pure whitening split
($\mathbf W_{\text{down}}=\mathbf G^{-1/2}\mathbf U_r$, $\mathbf W_{\text{up}}=\mathbf\Sigma_r \mathbf V_r^\top$) spans the same subspace but yields
a unit-variance latent, discarding the energy ordering; the square-root split is what couples the
transform to the bit allocator of Section~\ref{subsec:quant}.}

Two refinements complete the transform. First, the low-rank factorization of the value path leaves
a small, structured residual at the attention output; since the output projection $\mathbf W_O$ lies outside
the cache, we absorb the dominant part of this residual with a closed-form, training-free reduced-rank
correction to $\mathbf W_O$ (Appendix~\ref{app:healing}), which changes neither the cached object nor the cost
of inference. Second, the rank budget $B$ is split asymmetrically between the two paths,
$r_k = \kappa B$ and $r_v = (1-\kappa)B$: keys' produced energy concentrates faster than values',
so keys tolerate a smaller rank (more compression) and values receive the
larger share of the rank budget ($\kappa<\tfrac12$). Both factors $\mathbf W_{\text{down}}, \mathbf W_{\text{up}}$ and the $\mathbf W_O$
correction are folded back into the model weights, so the deployed model is a drop-in replacement of
$\mathbf W_K$ and $\mathbf W_V$ with no extra runtime structure.

Beyond shrinking the cache, the transform hands the next stage a
canonical, energy-ordered latent with measured coordinate means and
variances. This gives the bit allocator a principled axis of importance, which
we turn to next.

\subsection{Rate--Distortion Quantization of the Latent}
\label{subsec:quant}

Because the latent coordinates are energy-ordered, quantizing them all at one
bit-width is wasteful: a high-variance coordinate and a near-zero one would be
given the same precision despite contributing very differently to the
reconstruction. After subtracting the measured coordinate means, we model the
centered coordinates as independent Gaussian scalar sources with marginal
variances $v_j$. This independence assumption is an approximation, because the
$G$-weighted transform guarantees diagonal uncentered second moments, not
diagonal centered covariance. Under the scalar-source model, quantization at
$b_j$ bits incurs distortion proportional to $v_j\,2^{-2b_j}$ at high
rate~\citep{gersho1992vector,720541}, and minimizing
$\sum_j v_j 2^{-2b_j}$ subject to
$\tfrac1r\sum_j b_j=\bar b$ yields the classical \emph{reverse water-filling} allocation~\citep{1088759,cover2006elements} (Figure~\ref{fig:method}(b))
\begin{equation}
b_j = \bar b + \tfrac12\log_2\!\Big(v_j \,/\, \mathrm{GM}(v)\Big),
\qquad \mathrm{GM}(v)=\Big(\textstyle\prod_j v_j\Big)^{1/r},
\label{eq:waterfill}
\end{equation}
clamped to $[0,b_{\max}]$ (derivation in Appendix~\ref{app:waterfill}).
Half a bit is added for every doubling of variance above the geometric mean; the optimum equalizes
the residual distortion left in each retained coordinate, not the bits spent on it.

The transform supplies what the allocation needs operationally: a canonical
energy ordering and measured marginal variances. Raw $\mathbf K,\mathbf V$ channels are
correlated and unordered, while random rotations such as TurboQuant or
QuaRot~\citep{zandieh2025turboquantonlinevectorquantization,ashkboos2024quarotoutlierfree4bitinference}
impose no data-dependent ordering. Together, the $G$-weighted transform of
Section~\ref{subsec:latent} and the per-coordinate allocation here follow the
classical transform-coding pattern~\citep{952802}: transform a source into an
ordered coordinate system, then allocate rate across its coefficients.


Two properties make the allocation practical (Figure~\ref{fig:bitprofile}, Appendix~\ref{app:waterfill}).
First, a coordinate given zero bits is dropped entirely, so as $\bar b$ falls the low-energy tail is
zeroed and quantization passes \emph{continuously} into rank reduction: the effective rank is an output
of the allocator, not a separate hyperparameter. Second, because per-coordinate bit-widths do not
compose with a streaming cache, we group $g$ adjacent energy-ordered coordinates to share one bit-width
(from Equation~\eqref{eq:waterfill} via the group's mean variance) and quantize each group with per-token min/max
scales, keeping the cache append-only and outlier-robust (Figure~\ref{fig:method}(c)). Each group stores
its payload plus one fp16 scale and zero-point per token, both counted in the effective rate; zero-bit
groups store nothing.

\subsection{Operating Points and System Co-design}
\label{subsec:system}

\paragraph{From storage to compute.} Caching the latent also opens a compute path closed to scalar
quantizers: because the score is linear in the latent, the up-projection folds into the query and
attention runs directly in the $r$-dimensional latent space (Mode~B). All quality results, however, use
the reconstruct-then-attend path (Mode~A, Figure~\ref{fig:method}(c)): each token is down-projected,
quantized, and appended, and $\mathbf K,\mathbf V$ are rebuilt on the fly, which is RoPE-exact and
training-free. 

%% file: figs/fig_method.tex
\begin{figure}[t]
\centering
\includegraphics[width=\textwidth]{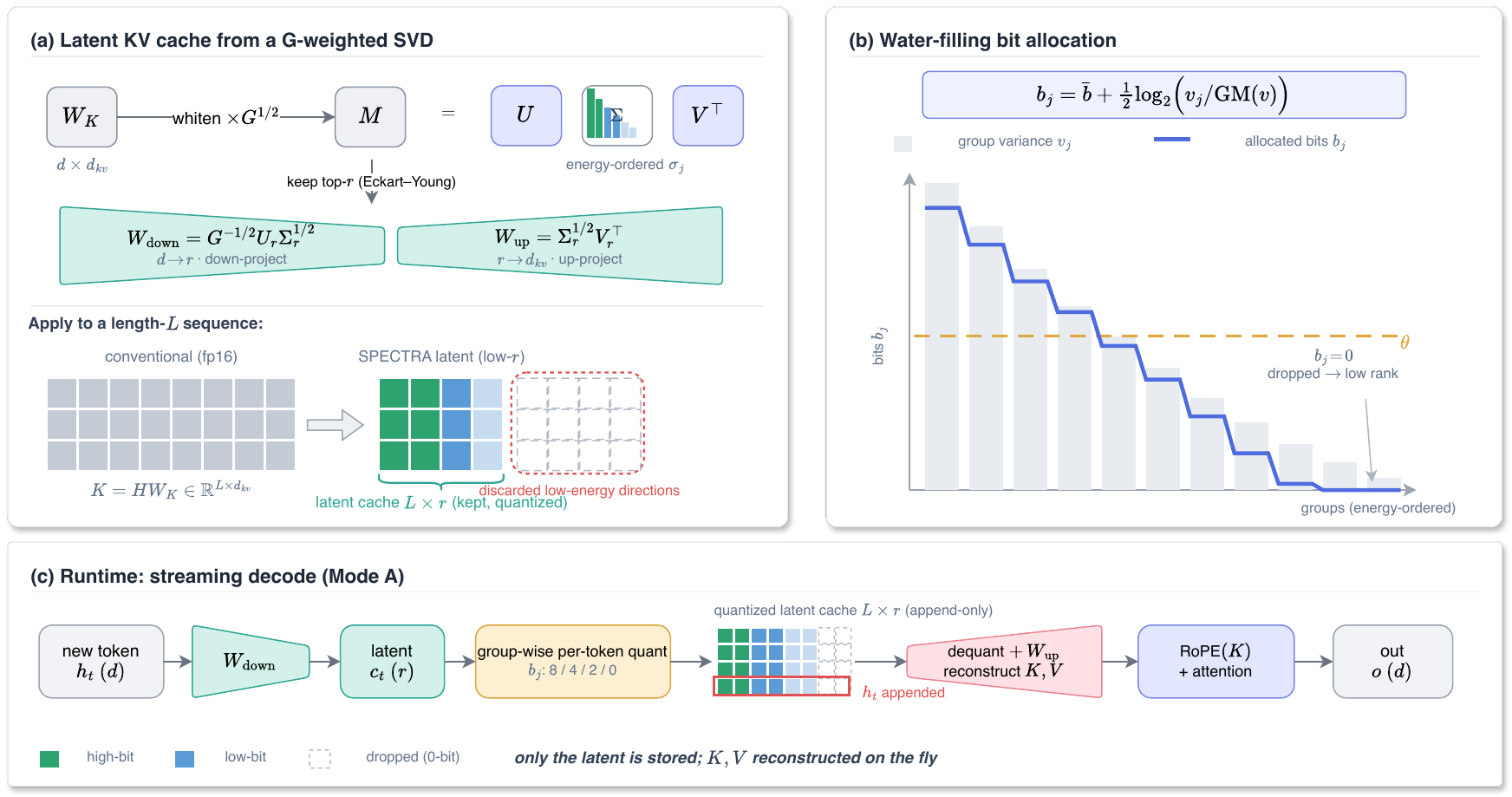}
\caption{\textbf{\spectra{} method overview.}
\textbf{(a) Latent KV cache from a $G$-weighted SVD.} Each key/value projection is factorized from the
whitened weight $M=G^{1/2}W$; keeping the top-$r$ singular directions (Eckart--Young) yields
$W_{\text{down}},W_{\text{up}}$, so a length-$L$ sequence is cached as an $L\times r$ energy-ordered
latent instead of the $L\times d_{kv}$ full-precision $K,V$; the low-energy directions are discarded by the low-rank truncation (rank reduction).
\textbf{(b) Water-filling bit allocation.} Reverse water-filling against a level $\theta$ assigns more
bits to high-variance latent groups and drives the low-energy tail to zero bits, i.e.\ rank reduction.
\textbf{(c) Streaming decode (Mode A).} Each new token is down-projected, quantized per token with the
group bit-widths, and appended to the latent cache; $K,V$ are reconstructed on the fly (RoPE applied to
$K$) for standard attention.}
\label{fig:method}
\end{figure}

%% file: sections/04_experiments.tex
\section{Experiments}
\label{sec:exp}
\subsection{Settings}

\paragraph{Models.}
We evaluate \spectra{} on Llama-3.1-8B-Instruct~\citep{grattafiori2024llama3herdmodels},
Mistral-7B-Instruct-v0.3~\citep{jiang2023mistral7b}, and
Qwen2.5-7B-Instruct~\citep{qwen2025qwen25technicalreport}. All three use \emph{grouped-query attention}
(GQA), an attention layout in which several query heads share a single key--value head: Llama and
Mistral have 32 query and 8 key--value heads (a 1024-dimensional per-token KV cache), while Qwen2.5-7B
has 28 query and 4 key--value heads (a 512-dimensional per-token KV cache). We cap contexts at 31,500
tokens so that all methods share a common evaluation protocol. \spectra{} is implemented on top of
Hugging Face Transformers and applied post hoc, without any fine-tuning.

\paragraph{Tasks.}
We evaluate on two long-context suites. \textbf{LongBench}~\citep{bai2024longbench} contributes eight
English tasks spanning single-document QA (Qasper), query-based and multi-document summarization
(QMSum, MultiNews), few-shot classification (TREC), few-shot QA (TriviaQA), dialogue
summarization (SAMSum), and repository-level code completion (LCC, RepoBench-P); we use the first
$150$ examples of each task. \textbf{Needle-in-a-Haystack} (NIAH), which hides a short target string in
a long distractor context and asks the model to retrieve it, measures long-context retrieval on a
grid of six context lengths ($1$k--$32$k) and seven needle depths, in both single-needle and
four-needle variants.

\paragraph{Evaluation protocol.}
We follow the official LongBench harness: official per-task prompt templates and metrics (token-level
$F_1$ for QA, ROUGE-L for summarization, classification accuracy for TREC, and edit similarity for
code), middle-truncation of inputs longer than $31{,}500$ tokens (equal-length prefix and suffix), the
chat template applied to all tasks except the few-shot and code tasks
(TREC/TriviaQA/SAMSum/LCC/RepoBench-P, which the official protocol feeds without a chat wrapper), and
greedy decoding with the task-specific generation limits. Every method including dense, \spectra{}, and all
baselines, is run through this \emph{identical} harness, so the comparison is strictly
apples-to-apples. NIAH scores the fraction of planted magic codes recovered by greedy generation.

\paragraph{Baselines.}
We compare against five KV-cache compression methods, each run from its official implementation but
evaluated under the harness above: KIVI~\citep{liu2024kivi},
TurboQuant~\citep{zandieh2025turboquantonlinevectorquantization},
PolarQuant~\citep{han2025polarquantquantizingkvcaches},
OTT~\citep{su2025accuratekvcachequantization}, and
RotateKV~\citep{su2025rotatekvaccuraterobust2bit}. Baselines are run on Llama-3.1-8B and Mistral-7B;
for Qwen2.5-7B we report \spectra{} against the dense model only. For every method we report the best
configuration per compression tier, and all reported ratios are effective, counting the metadata each
method stores; the full sweep and configuration details are in Appendix~\ref{app:full_results}.

\subsection{Overall Results}
\label{subsec:overall}

\begin{figure}[ht]
\centering
\includegraphics[width=0.72\linewidth]{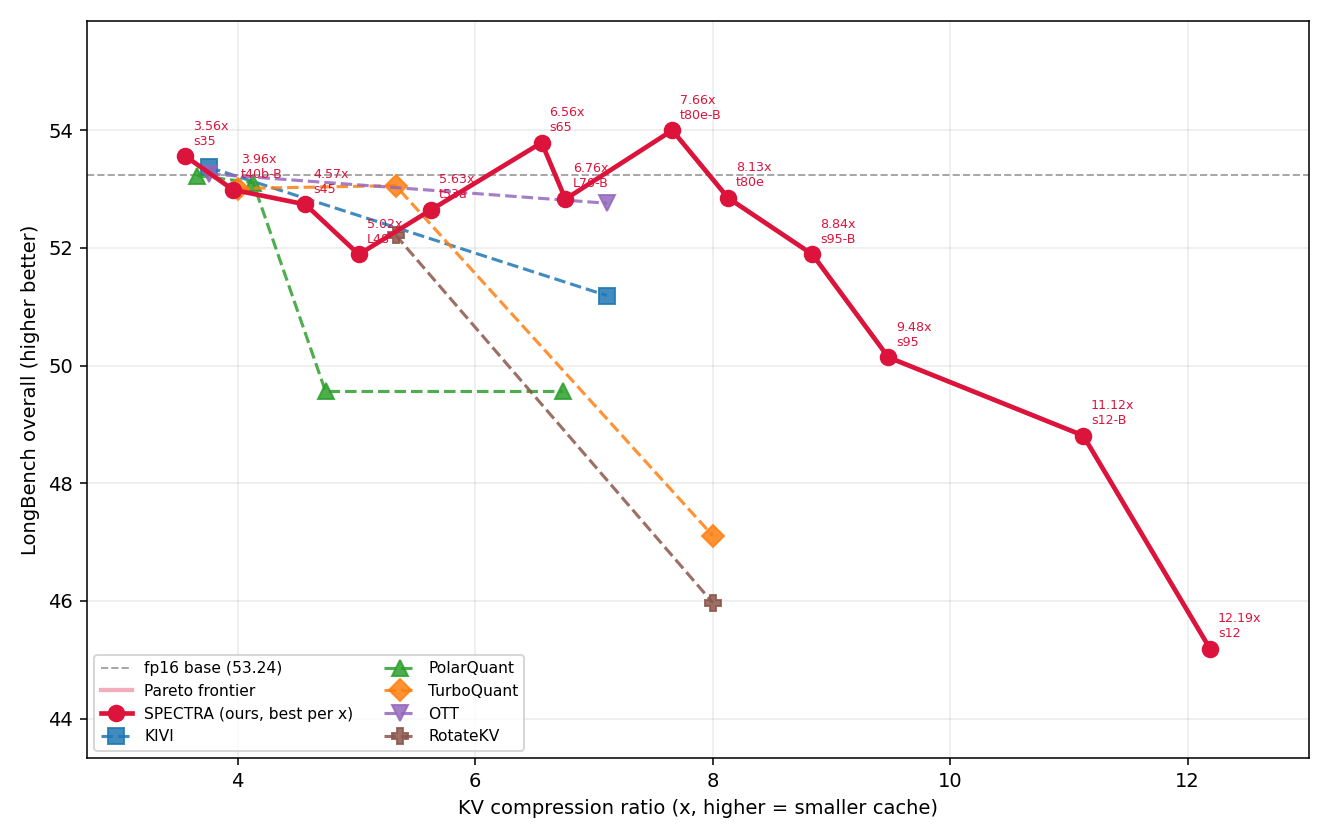}
\caption{\textbf{LongBench overall vs.\ KV-cache compression --- Llama-3.1-8B-Instruct.}
Average LongBench score (8 EN tasks) as a function of the effective KV compression ratio
(higher $x$ = smaller cache). SPECTRA (red) forms a single Pareto frontier that stays within
$\sim$1.5 points of the fp16 base (53.24, dashed) out to $\sim$8$\times$ and remains usable at
11.12$\times$ (48.81), while the scalar/rotation baselines (KIVI, OTT, TurboQuant, RotateKV, PolarQuant)
fall off the 2-bit cliff by $\sim$8$\times$ (TurboQuant 47.11, RotateKV 45.97) and none reach past
8$\times$. SPECTRA is on or above the frontier across the entire range; point labels give SPECTRA's
compression ratio and best config per tier.}
\label{fig:llama_lb_curve}
\end{figure}

Tables~\ref{tab:lb-llama3.1-8b} and~\ref{tab:lb-mistral-7b} report LongBench, and
Figure~\ref{fig:llama_lb_curve} plots the resulting quality--compression frontier. Two things stand
out in our experiment result: at low compression \spectra{} is effectively lossless and competitive with the strongest
baselines: at $3.5\times$ it scores $53.56$ on Llama and $53.50$ on Mistral, above the dense model and ahead of KIVI, OTT, and PolarQuant in the same range of compression rate. The decisive gap is at high
compression. By approximately $8\times$ the scalar and rotation quantization method fall off the $2$-bit cliff of
Section~\ref{sec:motivation} --- on Llama, TurboQuant and RotateKV drop to $47.11$ and $45.97$ from a
dense $53.24$, and none of the five baselines report a usable point past $8\times$. \spectra{} does not
collapse: it holds $54.00$ at $7.7\times$ and $51.89$ at $8.84\times$ compression, and even stays usable out to $11\times$ ($48.8$), tracing a single frontier no baseline reaches. Retrieval behaves the same way --- \spectra{}
stays lossless on single-needle NIAH out to $12\times$, where the quantization method degrade (Appendix~\ref{app:full_results_niah}).

\begin{table}[H]\centering\small\setlength{\tabcolsep}{3.5pt}
\caption{\textbf{LongBench per task --- Llama-3.1-8B-Instruct.} One SPECTRA row per compression tier for head-to-head with baselines, plus SPECTRA's high-compression points ($>$8$\times$) that no baseline reaches. Within each tier the \colorbox{blue!18}{\textbf{best}} and \colorbox{blue!7}{\underline{second}} Avg are highlighted.}
\label{tab:lb-llama3.1-8b}\scalebox{0.80}{%
\begin{tabular}{l c cccccccc c}\toprule
Method & Comp. & Qasper & QMSum & MNews & TREC & TQA & SAMSum & LCC & RB-P & Avg \\\midrule
base (fp16) & $1.00\times$ & 46.43 & 25.19 & 26.66 & 70.67 & 91.21 & 42.97 & 65.68 & 57.08 & \textbf{53.24} \\\midrule
\cellcolor{blue!18}\textbf{SPECTRA} & $3.56\times$ & 47.25 & 25.22 & 26.58 & 70.67 & 91.13 & 43.18 & 65.70 & 58.72 & \cellcolor{blue!18}\textbf{53.56} \\
PolarQuant & $3.66\times$ & 46.07 & 25.07 & 26.51 & 70.67 & 91.20 & 42.87 & 65.31 & 58.11 & 53.23 \\
OTT & $3.76\times$ & 46.32 & 25.26 & 26.33 & 70.67 & 90.99 & 43.20 & 65.22 & 58.09 & 53.26 \\
\cellcolor{blue!7}\underline{KIVI} & $3.76\times$ & 45.84 & 25.15 & 26.75 & 70.67 & 91.32 & 43.89 & 65.27 & 58.05 & \cellcolor{blue!7}\underline{53.37} \\
TurboQuant & $4.00\times$ & 45.41 & 25.28 & 26.26 & 72.00 & 89.38 & 43.52 & 65.95 & 56.27 & 53.01 \\
PolarQuant & $4.13\times$ & 45.87 & 25.21 & 26.86 & 70.67 & 91.31 & 42.12 & 65.63 & 57.19 & 53.11 \\
\midrule
\cellcolor{blue!7}\underline{SPECTRA} & $4.57\times$ & 45.70 & 25.36 & 27.13 & 70.00 & 90.41 & 43.30 & 64.60 & 55.39 & \cellcolor{blue!7}\underline{52.74} \\
PolarQuant & $4.74\times$ & 43.10 & 24.55 & 25.39 & 63.33 & 90.82 & 39.61 & 60.44 & 49.24 & 49.56 \\
\cellcolor{blue!18}\textbf{TurboQuant} & $5.33\times$ & 45.33 & 25.18 & 26.12 & 70.00 & 90.84 & 42.30 & 66.55 & 58.14 & \cellcolor{blue!18}\textbf{53.06} \\
RotateKV & $5.33\times$ & 47.28 & 24.85 & 26.68 & 71.33 & 89.63 & 42.31 & 64.78 & 50.92 & 52.22 \\
\midrule
\cellcolor{blue!18}\textbf{SPECTRA} & $6.56\times$ & 43.44 & 24.71 & 25.79 & 70.00 & 91.47 & 42.63 & 66.28 & 66.03 & \cellcolor{blue!18}\textbf{53.79} \\
PolarQuant & $6.74\times$ & 43.10 & 24.55 & 25.39 & 63.33 & 90.82 & 39.61 & 60.44 & 49.24 & 49.56 \\
KIVI & $7.11\times$ & 41.62 & 24.38 & 26.30 & 69.33 & 90.61 & 42.89 & 61.88 & 52.49 & 51.19 \\
\cellcolor{blue!7}\underline{OTT} & $7.11\times$ & 44.30 & 24.88 & 26.57 & 70.67 & 90.90 & 44.14 & 64.35 & 56.25 & \cellcolor{blue!7}\underline{52.76} \\
\midrule
\cellcolor{blue!18}\textbf{SPECTRA} & $7.66\times$ & 47.23 & 25.45 & 25.64 & 74.00 & 88.67 & 42.26 & 65.39 & 63.39 & \cellcolor{blue!18}\textbf{54.00} \\
RotateKV & $8.00\times$ & 36.09 & 23.59 & 26.41 & 56.67 & 87.27 & 39.58 & 54.05 & 44.08 & 45.97 \\
\cellcolor{blue!7}\underline{TurboQuant} & $8.00\times$ & 42.19 & 24.02 & 23.93 & 58.67 & 88.02 & 43.25 & 52.72 & 44.07 & \cellcolor{blue!7}\underline{47.11} \\
\midrule
\cellcolor{blue!18}\textbf{SPECTRA} & $8.84\times$ & 43.77 & 24.94 & 25.56 & 70.00 & 85.49 & 42.14 & 62.79 & 60.41 & \cellcolor{blue!18}\textbf{51.89} \\
\midrule
\cellcolor{blue!18}\textbf{SPECTRA} & $11.12\times$ & 40.02 & 23.86 & 24.88 & 70.00 & 89.03 & 42.20 & 52.83 & 47.67 & \cellcolor{blue!18}\textbf{48.81} \\
\bottomrule\end{tabular}}\end{table}

\begin{table}[H]\centering\small\setlength{\tabcolsep}{3.5pt}
\caption{\textbf{LongBench per task --- Mistral-7B-Instruct-v0.3.} One SPECTRA row per compression tier for head-to-head with baselines, plus SPECTRA's high-compression points ($>$8$\times$) that no baseline reaches. Within each tier the \colorbox{blue!18}{\textbf{best}} and \colorbox{blue!7}{\underline{second}} Avg are highlighted.}
\label{tab:lb-mistral-7b}\scalebox{0.80}{%
\begin{tabular}{l c cccccccc c}\toprule
Method & Comp. & Qasper & QMSum & MNews & TREC & TQA & SAMSum & LCC & RB-P & Avg \\\midrule
base (fp16) & $1.00\times$ & 36.93 & 25.95 & 26.25 & 74.00 & 89.66 & 46.52 & 64.48 & 61.77 & \textbf{53.20} \\\midrule
\cellcolor{blue!18}\textbf{SPECTRA} & $3.56\times$ & 37.76 & 25.66 & 25.86 & 74.67 & 89.99 & 46.04 & 65.04 & 63.01 & \cellcolor{blue!18}\textbf{53.50} \\
OTT & $3.76\times$ & 36.95 & 25.78 & 26.19 & 74.00 & 90.06 & 46.41 & 62.79 & 60.67 & 52.86 \\
KIVI & $3.76\times$ & 35.90 & 25.75 & 26.18 & 74.00 & 90.06 & 46.37 & 62.99 & 59.44 & 52.59 \\
RotateKV & $4.00\times$ & 39.56 & 25.47 & 26.53 & 74.67 & 90.43 & 46.26 & 61.09 & 57.86 & 52.73 \\
\cellcolor{blue!7}\underline{TurboQuant} & $4.00\times$ & 37.66 & 25.36 & 26.35 & 74.00 & 89.73 & 46.05 & 63.39 & 61.71 & \cellcolor{blue!7}\underline{53.03} \\
\midrule
\cellcolor{blue!18}\textbf{SPECTRA} & $4.88\times$ & 37.74 & 26.00 & 26.05 & 74.67 & 90.23 & 44.61 & 63.77 & 61.06 & \cellcolor{blue!18}\textbf{53.02} \\
\cellcolor{blue!7}\underline{TurboQuant} & $5.33\times$ & 35.92 & 25.60 & 26.03 & 75.33 & 91.61 & 46.11 & 58.75 & 59.03 & \cellcolor{blue!7}\underline{52.30} \\
RotateKV & $5.33\times$ & 34.48 & 25.19 & 26.37 & 75.33 & 89.06 & 45.07 & 62.53 & 59.69 & 52.21 \\
\midrule
\cellcolor{blue!18}\textbf{SPECTRA} & $6.31\times$ & 37.70 & 25.93 & 25.82 & 75.33 & 89.70 & 44.51 & 62.79 & 61.78 & \cellcolor{blue!18}\textbf{52.94} \\
KIVI & $7.11\times$ & 36.09 & 24.94 & 25.67 & 74.00 & 89.84 & 46.17 & 61.60 & 56.36 & 51.83 \\
\cellcolor{blue!7}\underline{OTT} & $7.11\times$ & 35.64 & 24.68 & 26.16 & 74.00 & 90.01 & 46.28 & 62.63 & 58.47 & \cellcolor{blue!7}\underline{52.23} \\
\midrule
\cellcolor{blue!18}\textbf{SPECTRA} & $7.73\times$ & 36.29 & 25.95 & 25.97 & 72.00 & 89.41 & 45.56 & 58.38 & 55.81 & \cellcolor{blue!18}\textbf{51.17} \\
\cellcolor{blue!7}\underline{RotateKV} & $8.00\times$ & 30.24 & 24.79 & 26.25 & 70.00 & 87.31 & 42.88 & 59.87 & 53.64 & \cellcolor{blue!7}\underline{49.37} \\
TurboQuant & $8.00\times$ & 34.54 & 22.92 & 25.53 & 68.67 & 87.94 & 44.40 & 54.62 & 50.23 & 48.61 \\
\midrule
\cellcolor{blue!18}\textbf{SPECTRA} & $8.93\times$ & 32.59 & 25.84 & 25.99 & 74.00 & 88.39 & 45.56 & 56.85 & 54.05 & \cellcolor{blue!18}\textbf{50.41} \\
\midrule
\cellcolor{blue!18}\textbf{SPECTRA} & $11.28\times$ & 33.61 & 24.96 & 25.28 & 70.00 & 87.87 & 44.42 & 52.61 & 52.03 & \cellcolor{blue!18}\textbf{48.85} \\
\bottomrule\end{tabular}}\end{table}

\begin{figure}[ht]
\centering
\includegraphics[width=0.72\linewidth]{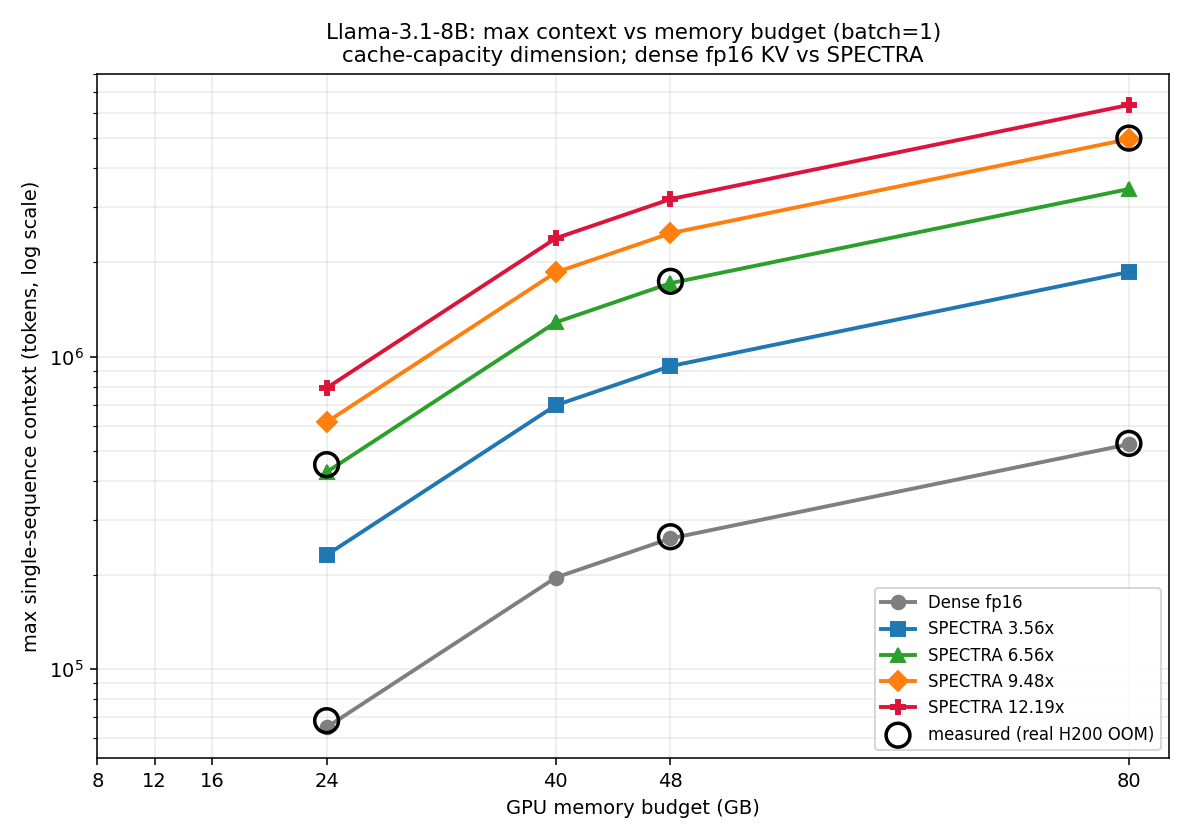}
\caption{\textbf{More context on the same hardware (Llama-3.1-8B, batch~1).} Longest single-sequence
context that fits before out-of-memory versus GPU memory budget, for the dense fp16 KV cache and
\spectra{} at representative compression ratios. Storing a compressed latent multiplies the context
that fits in a fixed budget; measured H200 OOM points match the analytical curve within $1\%$
(Table~\ref{tab:oom}). This isolates cache capacity, not an optimized attention kernel.}
\label{fig:oom}
\end{figure}

\subsection{Why cache a transformed latent?}

\begin{figure}[ht]
\centering
\includegraphics[width=0.72\linewidth]{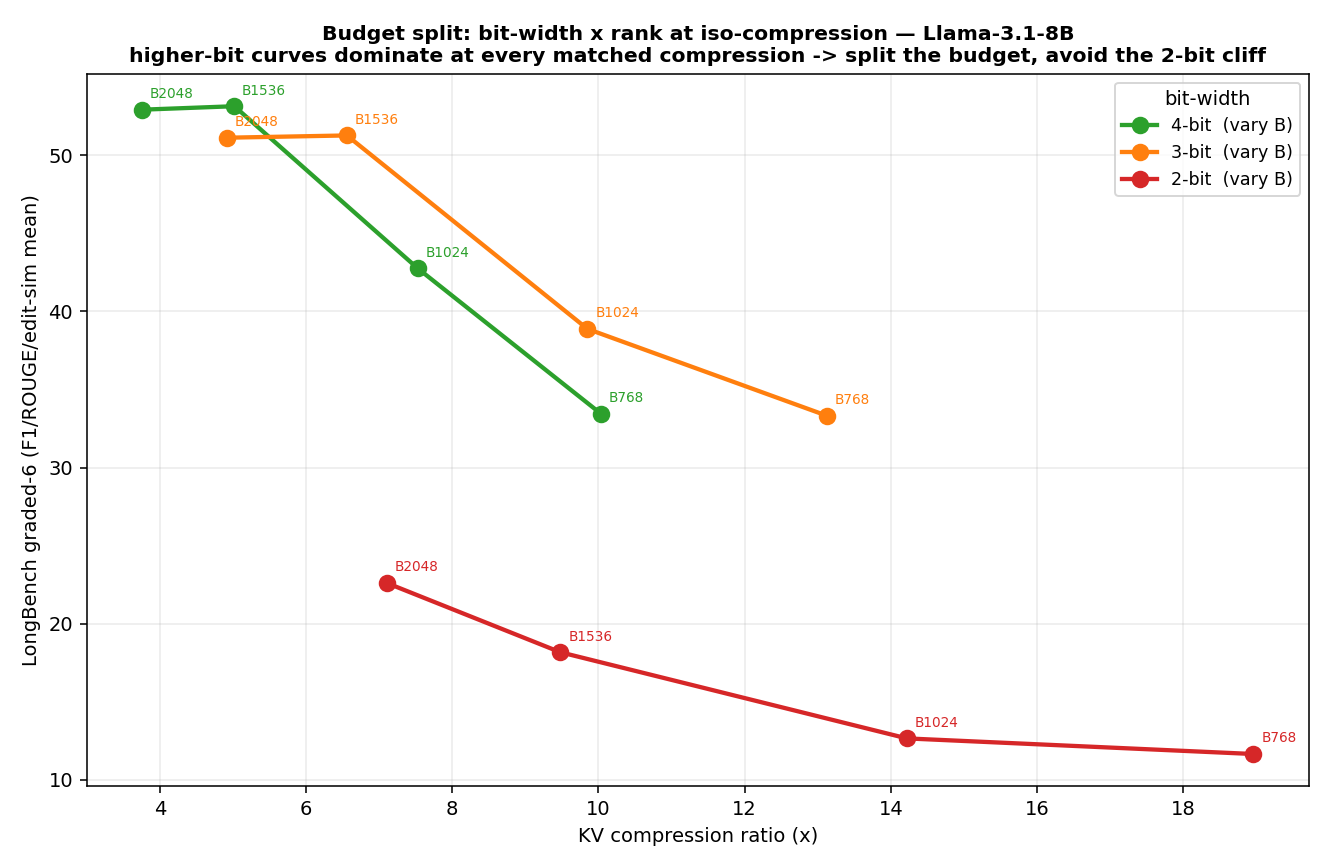}
\caption{\textbf{Splitting the budget beats spending it all on bits} (Llama-3.1-8B, $200$ samples).
LongBench (mean over six graded tasks) against effective compression, sweeping the latent width $B$
within each bit-width. At every matched compression the $3$- and $4$-bit curves dominate the $2$-bit
curve: given a fixed storage budget, distributing it across two axes---more bits \emph{and} more rank
reduction---is far better than driving a single axis to the $2$-bit cliff (red). \spectra{} therefore
never operates on the cliff; it moves along the upper envelope by trading rank for precision.}
\label{fig:ablation}
\end{figure}

We cache a \emph{transformed} latent rather than the raw cache, because the transform concentrates the
produced KV energy into a few ordered directions (Section~\ref{sec:motivation}), and that concentration
is what lets \spectra{} spend its budget on two axes at once instead of one. Figure~\ref{fig:ablation}
makes the consequence concrete: holding compression fixed and sweeping the latent width, the $3$- and
$4$-bit curves stay far above the $2$-bit curve everywhere, so a given storage budget goes much further
when it keeps more precision \emph{and} drops more rank than when it forces a single axis down to two
bits. \spectra{} therefore never has to sit on the $2$-bit cliff. Since rank reduction and quantization
are the two ends of one control (Section~\ref{sec:method}), it moves along the upper envelope of these
curves, trading rank for precision as the target tightens; the crossover where $3$ bits overtakes $4$
at high compression is that trade-off at work, since once the budget is tight, buying back rank at
slightly lower precision beats holding precision and cutting rank.

This compression translates directly into serving capacity. Figure~\ref{fig:oom} sweeps the GPU memory
budget and reports the longest single sequence that fits before running out of memory: on $24$\,GB the
dense fp16 cache caps a Llama-3.1-8B sequence at $65$k tokens, whereas \spectra{} fits $793$k at
$12\times$, and the measured H200 out-of-memory points track the analytical footprint within $1\%$
(Table~\ref{tab:oom}). We report capacity, not wall-clock speed: the latent-space decode path of
Section~\ref{sec:method} turns this saved memory into saved compute in principle, but we validate it
only with an isolated microbenchmark and leave end-to-end kernel integration to future work.

%% file: sections/02_related.tex
\section{Related Work}
\label{sec:related}

Two families dominate KV-cache compression. \emph{Quantization} lowers the bit-width of the stored
$K,V$, either directly~\citep{liu2024kivi,hooper2024kvquant} or after a decorrelating
rotation~\citep{ashkboos2024quarotoutlierfree4bitinference,liu2024spinquant,zandieh2025turboquantonlinevectorquantization,han2025polarquantquantizingkvcaches},
and degrades sharply near $2$ bits. \emph{Token eviction} keeps only salient
tokens~\citep{zhang2023h2oheavyhitteroracleefficient,li2024snapkvllmknowslooking,xiao2024streamingllm},
which is lossy on aggregation and multi-fact queries. Both fix the cached object to raw $K,V$;
\spectra{} is orthogonal and changes that object instead.

Closest to us are low-rank methods that cache an SVD-based
latent~\citep{chang2024palucompressingkvcachelowrank,zhang2024lorc} (with related weight low-rank
compression~\citep{wang2024svdllm,liu2026eorafinetuningfreecompensationcompressed}) and
latent-attention architectures that build a compressed latent into the
model~\citep{deepseekv2,meng2025transmla,ji2025mha2mla}. We differ in two
ways: (i) we factorize in the metric of the produced activations (a
$G$-weighted, KLT-like transform) rather than in weight space, retaining more
produced energy at equal rank and yielding a latent that is second-moment-orthogonal and energy-ordered; and (ii) we are post-hoc and training-free, unlike architectures that require pretraining or distillation. This latent makes \spectra{} an instance of classical transform coding, a Karhunen--Lo\`eve transform followed by rate--distortion (water-filling) bit allocation~\citep{952802,cover2006elements}, applied to the cache: unlike allocation over the token or layer axis, we allocate over the channels of a data-optimal transform. A fuller treatment with additional baselines and background appears in Appendix~\ref{app:extended_related}.

%% file: sections/05_conclusion.tex
\section{Conclusion}
\label{sec:conclusion}

We have argued that the main lever in KV-cache compression is not how finely to quantize the keys and
values, but what to store in the first place. The keys and values a model produces are highly
redundant, yet that redundancy becomes usable only once they are rotated into a coordinate system
computed from the model's own statistics, where a few channels carry almost all of the information and
the budget can be spent where it matters. \spectra{} turns this into a training-free, drop-in codec
that re-encodes the cache into this basis and gives each channel a bit-width that matches its
contribution, from several bits down to zero, so that low-rank reduction and quantization become two
settings of one control rather than separate tools. Because the transform folds into the key and value
projections, the model runs unchanged.

This lets \spectra{} push usable compression past the point where quantization alone collapses: it is
near-lossless at $4\times$, stays competitive at $8\times$ where uniform quantization has already
fallen off the $2$-bit cliff, and remains usable up to $12\times$, so the same GPU can hold far longer
contexts. How far the cache compresses tracks how concentrated its energy is, so the gains vary from
model to model, but they degrade gracefully rather than collapsing. Turning this saved memory into a
matching speedup is the natural next step: caching the latent already opens an attention path that runs
in the compressed space, and we leave a fused, end-to-end kernel for it to future work.

%% file: sections/appendix.tex
\section{Extended Related Work}
\label{app:extended_related}

\paragraph{KV-cache quantization.} The dominant approach to shrinking the cache lowers the bit-width
of the stored keys and values. KIVI~\citep{liu2024kivi} quantizes keys per-channel and values
per-token to $2$ bits; KVQuant~\citep{hooper2024kvquant} adds sensitivity-aware non-uniform
quantization and outlier handling. A second line first \emph{rotates} the cache to spread outliers
before quantizing, using random or learned orthogonal transforms
(QuaRot~\citep{ashkboos2024quarotoutlierfree4bitinference}, SpinQuant~\citep{liu2024spinquant},
TurboQuant~\citep{zandieh2025turboquantonlinevectorquantization}) or a polar reparameterization
(PolarQuant~\citep{han2025polarquantquantizingkvcaches}).
All of these keep the cached object fixed to raw $K,V$ and spend bits uniformly (or by heuristic
importance) over its channels; quality degrades sharply near $2$ bits (the ``$2$-bit cliff''). \spectra{}
instead quantizes a second-moment-orthogonal, energy-ordered latent, where
the measured marginal variances provide a canonical axis for rate allocation
and low-rank truncation carries compression that scalar quantizers can only
reach at INT2.

\paragraph{Token eviction and sparse attention.} A complementary family reduces the \emph{number} of
cached tokens rather than the bits per token, keeping a budget of salient entries.
H2O~\citep{zhang2023h2oheavyhitteroracleefficient} evicts by accumulated attention (heavy hitters),
SnapKV~\citep{li2024snapkvllmknowslooking} selects tokens using a lookahead window, and
StreamingLLM~\citep{xiao2024streamingllm} retains attention sinks plus a sliding window.
These methods are lossy on aggregation and multi-fact queries, where the discarded tokens are exactly
the evidence a later query needs. \spectra{} is orthogonal: it compresses \emph{every} token's
representation rather than dropping tokens, and could be composed with eviction.

\paragraph{Low-rank and latent KV compression.} Closest to us are methods that store a low-rank
projection of the cache. Palu~\citep{chang2024palucompressingkvcachelowrank} caches a low-rank latent
obtained from an SVD of the key/value projection weights; LoRC~\citep{zhang2024lorc} applies a
progressive low-rank compression across layers; SVD-LLM~\citep{wang2024svdllm} and
EoRA~\citep{liu2026eorafinetuningfreecompensationcompressed} use (truncation- or eigenspace-aware)
low-rank factorization for LLM weight compression.
The key difference is the \emph{metric}: these factorize in weight space, ranking directions by weight
energy, whereas \spectra{} minimizes error in the metric of the activations the model actually produces
($G$-weighted / KLT-like), which we show retains substantially more produced
energy at equal rank and, unlike a pure whitening or
random rotation, yields a latent that is both second-moment-orthogonal and
energy-ordered.

\paragraph{Latent attention as an architecture.} Multi-head latent attention
(MLA) in DeepSeek-V2~\citep{deepseekv2} caches a compressed latent by design, and recent work retrofits
existing models to MLA via architectural conversion and continued
training/distillation (TransMLA~\citep{meng2025transmla}, MHA2MLA~\citep{ji2025mha2mla}).
\spectra{} targets the same latent-cache idea but as a \emph{post-hoc, training-free} transform on an
already-trained model: no pretraining, fine-tuning, or distillation, and a drop-in replacement of the
key/value projections.

\paragraph{Transform coding and rate--distortion foundations.}
\spectra{} follows the classical transform-coding
pattern~\citep{952802,720541}: a data-dependent transform followed by bit
allocation over its coefficients. The classical KLT diagonalizes the centered
covariance of a source; our $G$-weighted transform instead diagonalizes the
uncentered latent second moment, after which we center the coefficients and
allocate using their measured marginal variances. Reverse water-filling is
rate--distortion optimal for parallel Gaussian
sources~\citep{1088759,cover2006elements,gersho1992vector}; the optimal low-rank factor in a weighted
metric follows from Eckart--Young~\citep{Eckart_Young_1936}, and our output-side correction is a
reduced-rank regression~\citep{IZENMAN1975248}. Our contribution is not these tools but their
application to the KV cache: identifying $G$ (the activation Gram matrix) as the right metric, and
showing that transform coding in that metric turns the ill-posed problem of quantizing raw $K,V$ into a
well-posed one.

\section{G-Weighted SVD}
\label{app:GSVD}

\paragraph{Activation-metric identity.} The same calculation underlies the objective of
Section~\ref{subsec:latent}. For any matrix $\Delta\mathbf W$ (e.g.\ the factorization residual
$\mathbf W - \mathbf W_{\text{down}}\mathbf W_{\text{up}}$), the expected error on produced activations is, with $\mathbf h$ a row vector,
\begin{equation}
\begin{aligned}
\mathbb{E}_{\mathbf h}\|\mathbf h\,\Delta\mathbf W\|^2
&= \mathbb{E}_{\mathbf h}\,\mathrm{tr}\!\big(\Delta\mathbf W^\top \mathbf h^\top \mathbf h\,\Delta\mathbf W\big) \\
&= \mathrm{tr}\!\big(\Delta\mathbf W^\top\, \mathbb{E}_{\mathbf h}[\mathbf h^\top \mathbf h]\,\Delta\mathbf W\big) \\
&= \mathrm{tr}\!\big(\Delta\mathbf W^\top \mathbf G\,\Delta\mathbf W\big) \\
&= \big\|\mathbf G^{1/2}\Delta\mathbf W\big\|_F^2 ,
\end{aligned}
\end{equation}
using linearity of the trace and expectation and the symmetry $\mathbf G^{1/2\top}\mathbf G^{1/2}=\mathbf G$. Minimizing this
over rank-$r$ $\mathbf W_{\text{down}}\mathbf W_{\text{up}}$ is thus a plain low-rank problem on the whitened weight
$\mathbf G^{1/2}\mathbf W$, solved in closed form by Equation~\ref{eq:gsvd}.

\paragraph{$G$-weighted (ours).} Let $\mathbf M=\mathbf G^{1/2}\mathbf W=\mathbf U\mathbf\Sigma \mathbf V^\top$. Truncating $\mathbf M$ to rank $r$ is
optimal in produced energy (Eckart--Young~\citep{Eckart_Young_1936} in the $G$-metric), so the retained fraction is
$\sum_{j\le r}\sigma_j^2 / \sum_j \sigma_j^2$, where $\sigma_j$ are the singular values of $\mathbf M$.

\paragraph{Plain SVD (Palu).} Let $\mathbf W=\mathbf U_w\mathbf\Sigma_w \mathbf V_w^\top$ be the SVD of the weight itself, and keep
its top-$r$ left singular directions. Scored in the \emph{same} produced-energy metric, the retained
fraction is
\begin{equation}
\frac{\sum_{j\le r} \sigma_{w,j}^2\,\big(\mathbf u_{w,j}^\top \mathbf G\, \mathbf u_{w,j}\big)}{\|\mathbf G^{1/2}\mathbf W\|_F^2},
\end{equation}
which follows from $\|\mathbf G^{1/2}\mathbf W_r\|_F^2=\sum_{j\le r}\sigma_{w,j}^2\,(\mathbf u_{w,j}^\top \mathbf G\,\mathbf u_{w,j})$ since the
right singular vectors are orthonormal. Because this basis orders directions by weight energy
$\sigma_{w,j}^2$ rather than by produced energy, it cannot retain more produced
energy than the $G$-weighted optimum at equal rank and generally retains less.

\paragraph{Setup.} $\mathbf G$ (and the mean $\bar{\mathbf h}$) are accumulated in fp64 over $24$ calibration windows
of length $2048$ from WikiText-2 train, via a forward hook on each layer's attention input; curves are
averaged over all layers.

\section{Closed-form \texorpdfstring{$\mathbf W_O$}{WO} healing}
\label{app:healing}

The low-rank factorization of the value path perturbs the attention output, and the error reaches the
residual stream through the output projection $\mathbf W_O$. Because $\mathbf W_O$ is not stored in the cache, we can
compensate the dominant, structured part of this error by adding a small correction $\Delta\mathbf W$ to
$\mathbf W_O$, at no inference cost. Crucially, the correction is fit on the \emph{low-rank but
un-quantized} model, so it heals the low-rank error, not the quantization noise (which is handled
separately in Section~\ref{subsec:quant}).

\paragraph{Objective.} Let $\mathbf Y$ be the input to $\mathbf W_O$ (the post-attention context) under the low-rank
model and $\Delta \mathbf O = \mathbf O_{\text{orig}} - \mathbf O_{\text{lowrank}}$ the resulting output error, collected over
calibration tokens. We want $\mathbf Y(\mathbf W_O + \Delta\mathbf W)\approx \mathbf Y \mathbf W_O + \Delta\mathbf O$, i.e.\ $\mathbf Y\,\Delta\mathbf W \approx
\Delta\mathbf O$, with $\Delta\mathbf W$ constrained to rank $\rho$:
\begin{equation}
\min_{\operatorname{rank}\Delta\mathbf W \le \rho}\ \big\| \mathbf Y\,\Delta\mathbf W - \Delta\mathbf O \big\|_F^2 .
\end{equation}

\paragraph{Closed form.} Let $\mathbf A = \mathbb{E}[\mathbf Y^\top \mathbf Y]$ and $\mathbf B = \mathbb{E}[\mathbf Y^\top \Delta\mathbf O]$ (both
accumulated as streaming second moments, $\mathcal{O}(d^2)$ memory, no per-token buffers). The unconstrained
least-squares solution is $\mathbf A^{-1}\mathbf B$; the optimal rank-$\rho$ solution whitens, truncates, and
un-whitens~\citep{IZENMAN1975248}:
\begin{equation}
\mathbf M = \mathbf A^{-1/2} \mathbf B = \mathbf U\mathbf\Sigma \mathbf V^\top,\qquad
\mathbf M_\rho = \mathbf U_{:\rho}\mathbf\Sigma_{:\rho}\mathbf V_{:\rho}^\top,
\end{equation}
\begin{equation}
\Delta\mathbf W_\rho = \mathbf A^{-1/2} \mathbf M_\rho,\qquad
\mathbf W_O \leftarrow \mathbf W_O + \Delta\mathbf W_\rho ,
\end{equation}
which is the optimal reduced-rank regression in the $\mathbf A$-metric. The SVD is computed once per layer, so
sweeping $\rho$ is essentially free; in practice $\rho=4$ recovers most of the lost quality at
negligible parameter cost. The keys/values are not healed this way: the value error enters $\mathbf W_O$
linearly and is compensable there, whereas the key error acts through the softmax and is not.

\section{Reverse water-filling derivation}
\label{app:waterfill}

We derive the allocation of Equation~\ref{eq:waterfill}, following classical optimal bit
allocation~\citep{1088759,cover2006elements}. Under the scalar-source
approximation of Section~\ref{subsec:quant}, treat the $r$ centered latent
coordinates as independent sources with measured marginal variances
$v_1\ge\cdots\ge v_r$. Under a high-rate scalar quantizer, coordinate $j$
quantized at $b_j$ bits incurs distortion $D_j = c\,v_j\,2^{-2b_j}$
for a constant $c$ common to all coordinates. We minimize the total distortion at a fixed average rate:
\begin{equation}
\min_{b_1,\dots,b_r}\ \sum_{j=1}^r v_j\, 2^{-2b_j}
\quad\text{s.t.}\quad \frac1r\sum_{j=1}^r b_j = \bar b .
\end{equation}
With a multiplier $\lambda$ for the rate constraint, the Lagrangian is
$\mathcal{L}=\sum_j v_j 2^{-2b_j} + \lambda\big(\sum_j b_j - r\bar b\big)$, and
$\partial\mathcal{L}/\partial b_j = -2\ln 2\,v_j 2^{-2b_j} + \lambda = 0$ gives
\begin{equation}
v_j\, 2^{-2b_j} \;=\; \frac{\lambda}{2\ln 2} \;=\; \theta
\qquad\text{(a constant, the ``water level'')},
\end{equation}
so the optimum equalizes the residual distortion $D_j$ across all retained coordinates. Solving,
$b_j = \tfrac12\log_2(v_j/\theta)$. Imposing the rate constraint,
$\tfrac1r\sum_j \tfrac12\log_2(v_j/\theta) = \bar b$, i.e.\
$\log_2\theta = \tfrac1r\sum_j \log_2 v_j - 2\bar b = \log_2\mathrm{GM}(v) - 2\bar b$
with $\mathrm{GM}(v)=(\prod_j v_j)^{1/r}$. Substituting back,
\begin{equation}
b_j = \bar b + \tfrac12\log_2\!\big(v_j / \mathrm{GM}(v)\big),
\end{equation}
which is Equation~\ref{eq:waterfill}. The unconstrained $b_j$ can be negative for low-variance
coordinates; clamping to $[0,b_{\max}]$ (reverse water-filling) sets those to zero bits --- the
coordinate is dropped, which is the continuous bridge to rank reduction. The integer, group-level
realization (per-group mean variance, integer rounding with a rate-matching fix-up) is described in
Section~\ref{subsec:quant}.

\begin{figure}[H]
\centering
\includegraphics[width=0.6\linewidth]{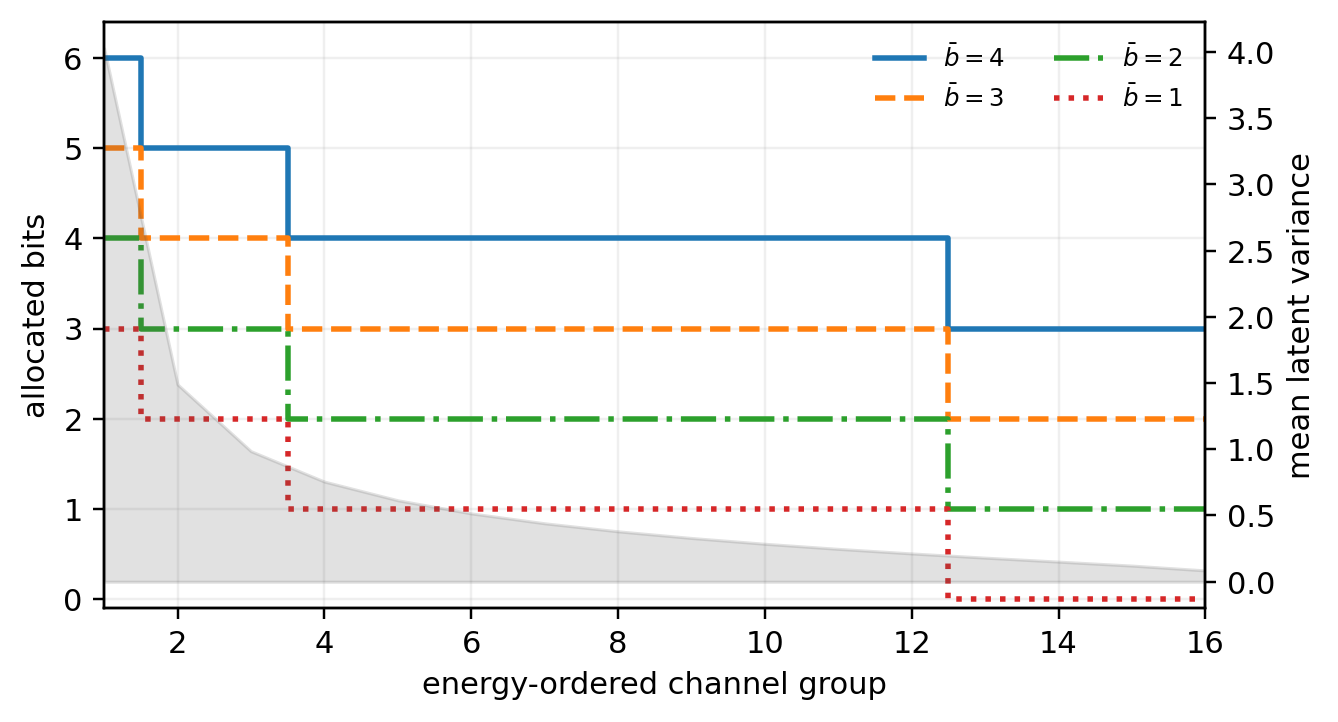}
\caption{\textbf{Reverse water-filling on a real layer} (Llama-3.1-8B, layer~16, keys). Allocated
bits (solid) track the latent energy spectrum (shaded): high-energy groups get more bits than the
uniform baseline (dotted, KIVI's implicit choice), low-energy groups fewer. As $\bar b{=}1$, tail
groups hit zero bits and quantization becomes rank reduction ($r_{\text{eff}}\approx768/1024$).}
\label{fig:bitprofile}
\end{figure}


\section{Latent-space attention and memory}
\label{app:system}

Caching the latent opens a path that is closed to scalar quantizers, because the attention score is
\emph{linear} in the latent:
\begin{equation}
\mathbf q\,\mathbf k_t^\top \;=\; \mathbf q\,(\mathbf c_t \mathbf W_{\text{up}})^\top \;=\; (\mathbf q\,\mathbf W_{\text{up}}^\top)\,\mathbf c_t^\top \;=\; \mathbf q'\,\mathbf c_t^\top,
\end{equation}
so the up-projection can be folded into the query once per decoding step ($\mathbf q'=\mathbf q\mathbf W_{\text{up}}^\top$,
a cheap $1\times d_{kv}\!\to\!1\times r$ map), and attention can then run \emph{directly} in the
$r$-dimensional latent space without ever materializing full keys or values. The value side is
$\mathrm{out}=\mathrm{softmax}(\mathbf q'\mathbf C^\top)\,\mathbf C_V\,\mathbf W_{\text{up}}^{V}$, where only the final $\cdot\,\mathbf W_{\text{up}}^{V}$
lifts back to $d_{kv}$. The dominant matrix multiplies ($\mathbf q'\mathbf C^\top$ and $\mathrm{softmax}\cdot \mathbf C_V$)
then act in dimension $r$ rather than $d_{kv}$, so the memory saving becomes a compute saving as well.
This is unavailable to KIVI/TurboQuant/PolarQuant, which cache $\mathbf k,\mathbf v$ directly and have no up-projection
to absorb.

\paragraph{From memory to compute (future work).} Because the dominant matrix multiplies act in
dimension $r$ rather than $d_{kv}$, the memory saving can in principle become a compute saving.
Realizing it as a wall-clock speedup, however, requires a fused latent-space attention kernel with
mixed-bit unpacking and RoPE integration---an engineering effort we deliberately leave to future work.
This paper therefore quantifies the dimension the latent already delivers \emph{without} new kernels:
memory.

\paragraph{Memory under constrained budgets.} A smaller per-token cache directly increases the context
(or batch) a fixed accelerator can hold. We sweep a range of GPU memory budgets and, for each, report
the longest single-sequence context that fits before out-of-memory, comparing the dense fp16 cache
against \spectra{} at representative compression ratios; the footprint used is the true compressed
size (latent payload plus per-group scales and zero-points), not an optimized runtime.
Figure~\ref{fig:oom} plots the result and Table~\ref{tab:oom} lists the analytical capacities. The
measured validation pins a filler tensor with the fp16 model resident and grows the compressed KV
footprint until CUDA OOM; measured H200 points match the analytical values within $1\%$ at every
tested budget and ratio (e.g.\ $24$\,GB dense $68.3$k measured vs.\ $68.8$k analytical;
$80$\,GB \spectra{}-$9.48\times$ $5.00$M measured vs.\ $5.00$M analytical).
\begin{table}[H]\centering\small\setlength{\tabcolsep}{6pt}
\caption{\textbf{Max single-sequence context (tokens) vs.\ GPU memory budget --- Llama-3.1-8B,
batch~1.} Geometry: $32$ layers, $8$ KV heads, head dim $128$ ($128$\,KiB/token dense fp16 KV); fp16
weights $16.06$\,GB. Storage counts the true compressed footprint (latent payload plus per-group
scales and zero-points). Real H200 validation tracks these values within $1\%$.}
\label{tab:oom}
\begin{tabular}{l ccccc}\toprule
Budget & Dense fp16 & \spectra{} $3.56\times$ & $6.56\times$ & $9.48\times$ & $12.19\times$ \\\midrule
$24$\,GB & $65$k & $232$k & $427$k & $617$k & $793$k \\
$40$\,GB & $196$k & $698$k & $1.29$M & $1.86$M & $2.39$M \\
$48$\,GB & $262$k & $931$k & $1.72$M & $2.48$M & $3.19$M \\
$80$\,GB & $524$k & $1.86$M & $3.44$M & $4.97$M & $6.39$M \\
\bottomrule\end{tabular}\end{table}

\section{Evaluation and compression-accounting details}
\label{app:eval-details}

\paragraph{Calibration.}
We estimate the activation Gram matrices and means from 2{,}048-token windows drawn from the
WikiText-2 training split, using $32$ windows for the LongBench and NIAH evaluations. Statistics are accumulated in fp64 from the input to each attention layer, and no evaluation examples are used to construct the transform. Unless noted otherwise, the latent group size is $g=64$ and the output correction has rank $\rho=4$; \spectra{}-WF uses an equal rank split between keys and values ($\kappa=0.5$), while \spectra{}-LR
assigns one quarter of the total rank budget to keys ($\kappa=0.25$).

\paragraph{Effective compression ratio.}
All reported compression ratios compare total KV-cache storage with the fp16
cache for the same model and sequence length. If $b_{\mathrm{eff}}$ denotes
the effective number of stored bits per original KV element, including the
payload, scales, and zero-points, then the reported ratio is
$16/b_{\mathrm{eff}}$. For \spectra{}-WF, $\bar b$ denotes the target average
payload rate before metadata rather than a uniform channel bit-width. For
\spectra{}-LR, the tables report the resulting effective ratio after both
rank reduction and latent quantization. Baselines are accounted the same way,
counting their per-group scales and zero-points, so all ratios are comparable.

\paragraph{Evaluation scope.}
LongBench uses the test split of \texttt{THUDM/LongBench} and the eight English tasks in
Tables~\ref{tab:lb-llama3.1-8b}--\ref{tab:lb-mistral-7b}, taking the first $150$ examples per task. We
use the official prompt templates and per-task metrics, truncate inputs longer than $31{,}500$ tokens
by keeping equal-length prefix and suffix segments, apply the chat template to all tasks except the
few-shot and code tasks (which the official protocol feeds without a wrapper), and decode greedily with
the task-specific generation limits. Needle-in-a-Haystack plants unique magic codes on a grid of six
context lengths ($1$k--$32$k) and seven depths, in single- and four-needle variants, and scores exact
recovery under greedy decoding. Every method---dense, \spectra{}, and all baselines---is evaluated
through this single harness. The memory experiment (Appendix~\ref{app:system}) sweeps a range of GPU
memory budgets and reports the longest single-sequence context that fits before out-of-memory, using
each method's true compressed cache footprint while the model stays resident; it isolates the
cache-capacity dimension and does not run an optimized compressed-attention kernel.

\section{Full Experiment Results}
\label{app:full_results}

\subsection{LongBench: Per-Task Results}
\label{app:full_results_longbench}
\input{sections/tab_llama_longbench_full}
\input{sections/tab_mistral_longbench_full}
\input{sections/tab_qwen_longbench}
\begin{figure}[H]
\centering
\includegraphics[width=0.7\linewidth]{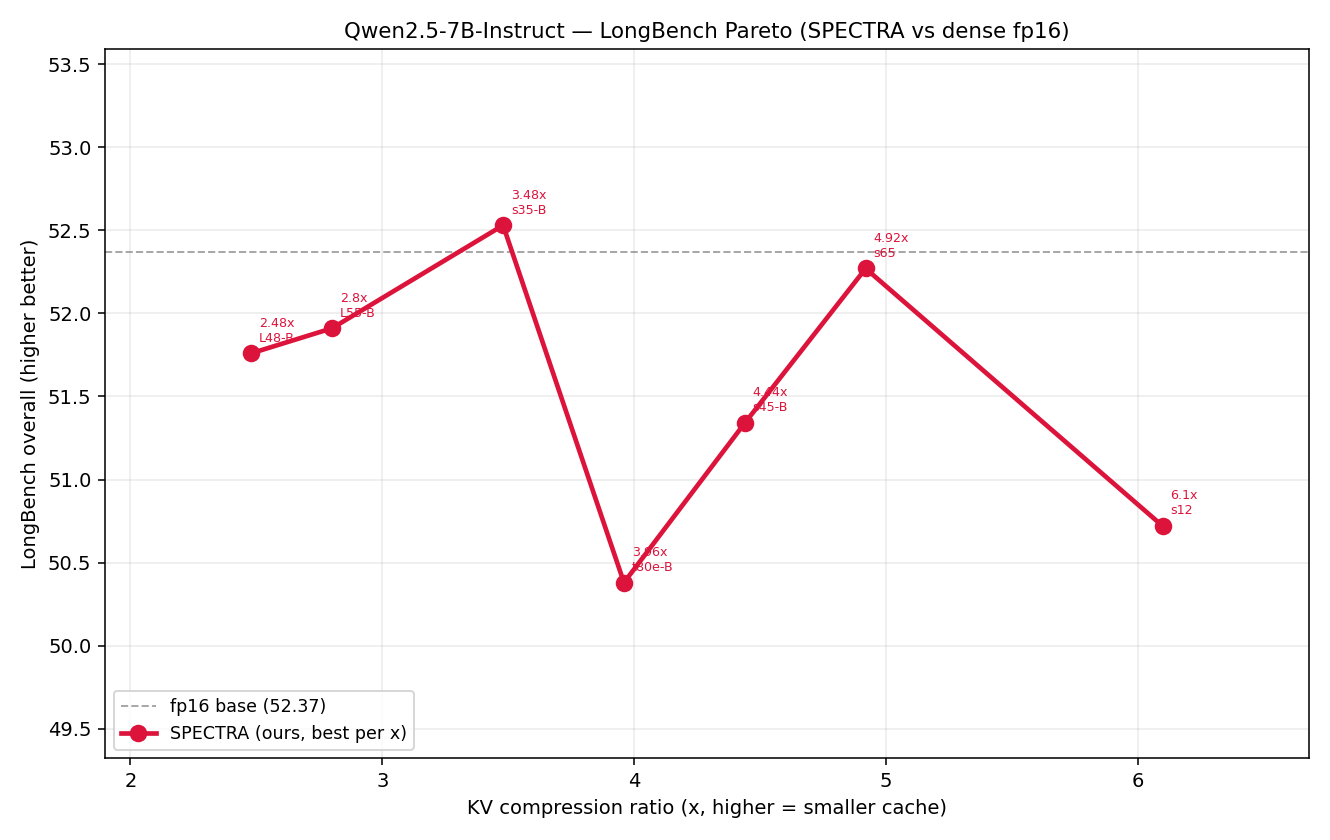}
\caption{\textbf{LongBench quality vs.\ compression --- Qwen2.5-7B-Instruct.} Dense line and \spectra{}
best-per-tier curve (no baselines run for this model). \spectra{} tracks the dense average out to
$\sim\!5\times$ and degrades beyond; the whole sweep tops out near $6\times$---lower than Llama's
$12\times$---consistent with Qwen's less concentrated produced-KV energy.}
\label{fig:qwen-pareto}
\end{figure}

\subsection{Needle-in-a-Haystack (NIAH)}
\label{app:full_results_niah}
\input{sections/fig_llama_niah}
\input{sections/fig_mistral_niah}
\begin{figure}[H]
\centering
\includegraphics[width=\linewidth]{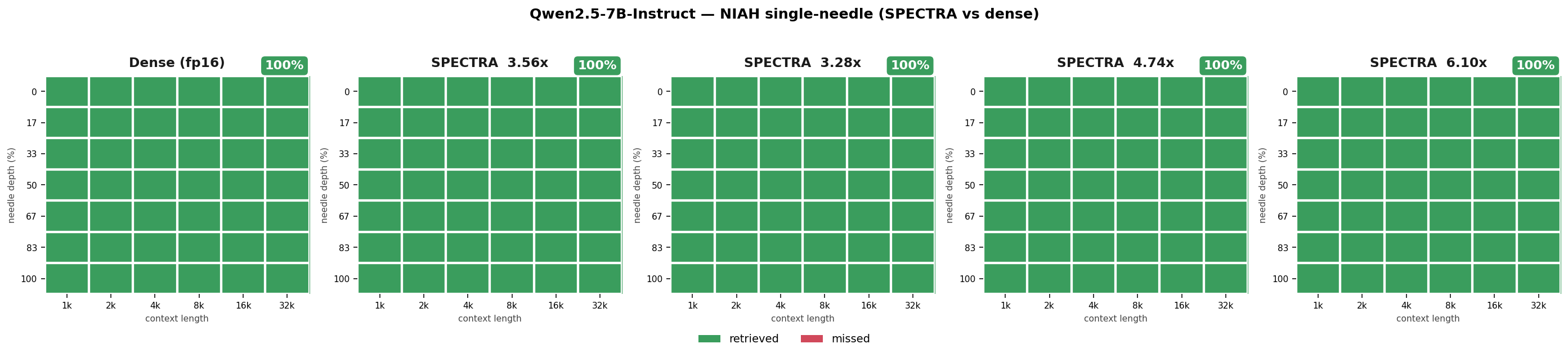}
\caption{\textbf{Needle-in-a-Haystack (single-needle) --- Qwen2.5-7B-Instruct.} Retrieval accuracy
over context length $\times$ depth for dense and \spectra{}; \spectra{} matches dense ($100\%$) at
every compression tier tested.}
\label{fig:qwen-niah}
\end{figure}
\begin{figure}[H]
\centering
\includegraphics[width=\linewidth]{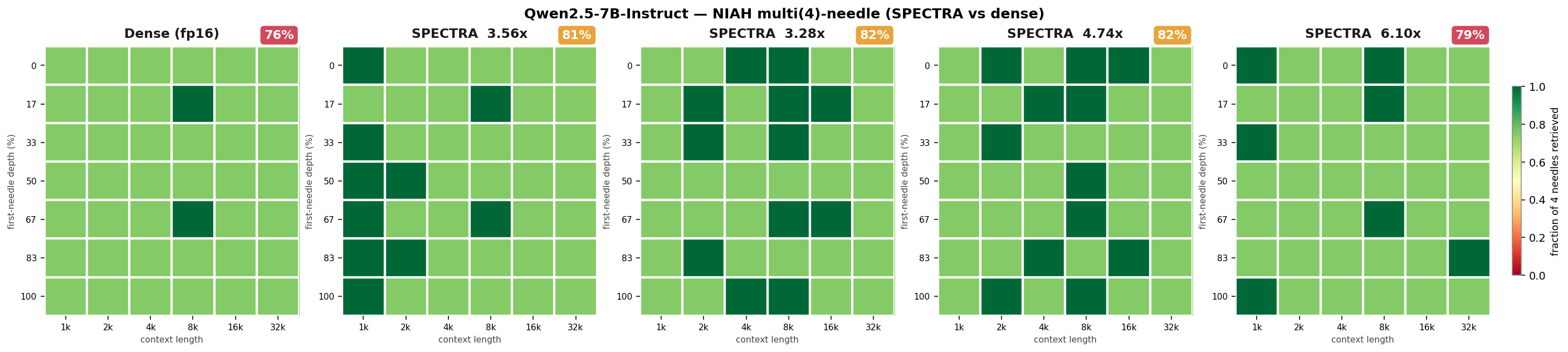}
\caption{\textbf{Needle-in-a-Haystack (multi-needle, 4) --- Qwen2.5-7B-Instruct.} \spectra{} retrieval
under the harder four-needle setting; it holds $79$--$82\%$, at or slightly above the dense model
($76\%$).}
\label{fig:qwen-niah-multi}
\end{figure}

%% file: sections/tab_llama_longbench_full.tex
\begin{table}[H]\centering\small\setlength{\tabcolsep}{3.5pt}
\caption{\textbf{LongBench per task --- Llama-3.1-8B-Instruct.} Full sweep: every SPECTRA operating point and all baselines, grouped by compression regime. Higher is better; Avg is the mean over the eight tasks. Within each group the \colorbox{blue!18}{\textbf{best}} and \colorbox{blue!7}{\underline{second}} Avg are highlighted.}
\label{tab:lb-llama3.1-8b-full}\resizebox{\textwidth}{!}{%
\begin{tabular}{l c cccccccc c}\toprule
Method & Comp. & Qasper & QMSum & MNews & TREC & TQA & SAMSum & LCC & RB-P & Avg \\\midrule
base (fp16) & $1.00\times$ & 46.43 & 25.19 & 26.66 & 70.67 & 91.21 & 42.97 & 65.68 & 57.08 & \textbf{53.24} \\\midrule
\cellcolor{blue!18}\textbf{SPECTRA} & $3.56\times$ & 47.25 & 25.22 & 26.58 & 70.67 & 91.13 & 43.18 & 65.70 & 58.72 & \cellcolor{blue!18}\textbf{53.56} \\
PolarQuant & $3.66\times$ & 46.07 & 25.07 & 26.51 & 70.67 & 91.20 & 42.87 & 65.31 & 58.11 & 53.23 \\
OTT & $3.76\times$ & 46.32 & 25.26 & 26.33 & 70.67 & 90.99 & 43.20 & 65.22 & 58.09 & 53.26 \\
\cellcolor{blue!7}\underline{KIVI} & $3.76\times$ & 45.84 & 25.15 & 26.75 & 70.67 & 91.32 & 43.89 & 65.27 & 58.05 & \cellcolor{blue!7}\underline{53.37} \\
SPECTRA & $3.96\times$ & 46.61 & 25.54 & 25.89 & 70.67 & 89.58 & 44.17 & 62.17 & 59.29 & 52.99 \\
TurboQuant & $4.00\times$ & 45.41 & 25.28 & 26.26 & 72.00 & 89.38 & 43.52 & 65.95 & 56.27 & 53.01 \\
PolarQuant & $4.13\times$ & 45.87 & 25.21 & 26.86 & 70.67 & 91.31 & 42.12 & 65.63 & 57.19 & 53.11 \\
\midrule
\cellcolor{blue!7}\underline{SPECTRA} & $4.57\times$ & 45.70 & 25.36 & 27.13 & 70.00 & 90.41 & 43.30 & 64.60 & 55.39 & \cellcolor{blue!7}\underline{52.74} \\
PolarQuant & $4.74\times$ & 43.10 & 24.55 & 25.39 & 63.33 & 90.82 & 39.61 & 60.44 & 49.24 & 49.56 \\
SPECTRA & $5.02\times$ & 43.29 & 24.84 & 26.40 & 69.33 & 89.73 & 44.61 & 62.30 & 54.61 & 51.89 \\
\cellcolor{blue!18}\textbf{TurboQuant} & $5.33\times$ & 45.33 & 25.18 & 26.12 & 70.00 & 90.84 & 42.30 & 66.55 & 58.14 & \cellcolor{blue!18}\textbf{53.06} \\
RotateKV & $5.33\times$ & 47.28 & 24.85 & 26.68 & 71.33 & 89.63 & 42.31 & 64.78 & 50.92 & 52.22 \\
SPECTRA & $5.63\times$ & 44.52 & 25.56 & 26.41 & 68.67 & 90.44 & 43.49 & 62.95 & 59.17 & 52.65 \\
\midrule
\cellcolor{blue!18}\textbf{SPECTRA} & $6.56\times$ & 43.44 & 24.71 & 25.79 & 70.00 & 91.47 & 42.63 & 66.28 & 66.03 & \cellcolor{blue!18}\textbf{53.79} \\
PolarQuant & $6.74\times$ & 43.10 & 24.55 & 25.39 & 63.33 & 90.82 & 39.61 & 60.44 & 49.24 & 49.56 \\
\cellcolor{blue!7}\underline{SPECTRA} & $6.76\times$ & 43.78 & 24.95 & 25.87 & 70.67 & 89.56 & 42.70 & 66.47 & 58.67 & \cellcolor{blue!7}\underline{52.83} \\
KIVI & $7.11\times$ & 41.62 & 24.38 & 26.30 & 69.33 & 90.61 & 42.89 & 61.88 & 52.49 & 51.19 \\
OTT & $7.11\times$ & 44.30 & 24.88 & 26.57 & 70.67 & 90.90 & 44.14 & 64.35 & 56.25 & 52.76 \\
\midrule
\cellcolor{blue!18}\textbf{SPECTRA} & $7.66\times$ & 47.23 & 25.45 & 25.64 & 74.00 & 88.67 & 42.26 & 65.39 & 63.39 & \cellcolor{blue!18}\textbf{54.00} \\
RotateKV & $8.00\times$ & 36.09 & 23.59 & 26.41 & 56.67 & 87.27 & 39.58 & 54.05 & 44.08 & 45.97 \\
TurboQuant & $8.00\times$ & 42.19 & 24.02 & 23.93 & 58.67 & 88.02 & 43.25 & 52.72 & 44.07 & 47.11 \\
\cellcolor{blue!7}\underline{SPECTRA} & $8.13\times$ & 40.11 & 25.77 & 25.29 & 69.33 & 89.34 & 42.38 & 65.44 & 65.13 & \cellcolor{blue!7}\underline{52.85} \\
\midrule
\cellcolor{blue!18}\textbf{SPECTRA} & $8.84\times$ & 43.77 & 24.94 & 25.56 & 70.00 & 85.49 & 42.14 & 62.79 & 60.41 & \cellcolor{blue!18}\textbf{51.89} \\
\cellcolor{blue!7}\underline{SPECTRA} & $9.48\times$ & 38.37 & 24.46 & 25.86 & 68.67 & 86.56 & 41.00 & 62.55 & 53.61 & \cellcolor{blue!7}\underline{50.14} \\
\midrule
\cellcolor{blue!18}\textbf{SPECTRA} & $11.12\times$ & 40.02 & 23.86 & 24.88 & 70.00 & 89.03 & 42.20 & 52.83 & 47.67 & \cellcolor{blue!18}\textbf{48.81} \\
\cellcolor{blue!7}\underline{SPECTRA} & $12.19\times$ & 31.54 & 23.59 & 25.12 & 65.33 & 86.11 & 41.85 & 45.33 & 42.59 & \cellcolor{blue!7}\underline{45.18} \\
\bottomrule\end{tabular}}\end{table}

%% file: sections/tab_mistral_longbench_full.tex
\begin{table}[H]\centering\small\setlength{\tabcolsep}{3.5pt}
\caption{\textbf{LongBench per task --- Mistral-7B-Instruct-v0.3.} Full sweep: every SPECTRA operating point and all baselines, grouped by compression regime. Higher is better; Avg is the mean over the eight tasks. Within each group the \colorbox{blue!18}{\textbf{best}} and \colorbox{blue!7}{\underline{second}} Avg are highlighted.}
\label{tab:lb-mistral7b-v0.3}\resizebox{\textwidth}{!}{%
\begin{tabular}{l c cccccccc c}\toprule
Method & Comp. & Qasper & QMSum & MNews & TREC & TQA & SAMSum & LCC & RB-P & Avg \\\midrule
base (fp16) & $1.00\times$ & 36.93 & 25.95 & 26.25 & 74.00 & 89.66 & 46.52 & 64.48 & 61.77 & \textbf{53.20} \\\midrule
\cellcolor{blue!18}\textbf{SPECTRA} & $3.56\times$ & 37.76 & 25.66 & 25.86 & 74.67 & 89.99 & 46.04 & 65.04 & 63.01 & \cellcolor{blue!18}\textbf{53.50} \\
OTT & $3.76\times$ & 36.95 & 25.78 & 26.19 & 74.00 & 90.06 & 46.41 & 62.79 & 60.67 & 52.86 \\
KIVI & $3.76\times$ & 35.90 & 25.75 & 26.18 & 74.00 & 90.06 & 46.37 & 62.99 & 59.44 & 52.59 \\
SPECTRA & $3.97\times$ & 35.88 & 25.29 & 25.84 & 73.33 & 90.12 & 46.36 & 63.87 & 59.01 & 52.46 \\
RotateKV & $4.00\times$ & 39.56 & 25.47 & 26.53 & 74.67 & 90.43 & 46.26 & 61.09 & 57.86 & 52.73 \\
\cellcolor{blue!7}\underline{TurboQuant} & $4.00\times$ & 37.66 & 25.36 & 26.35 & 74.00 & 89.73 & 46.05 & 63.39 & 61.71 & \cellcolor{blue!7}\underline{53.03} \\
\midrule
\cellcolor{blue!7}\underline{SPECTRA} & $4.57\times$ & 35.84 & 25.50 & 26.44 & 72.67 & 89.59 & 46.16 & 63.86 & 61.16 & \cellcolor{blue!7}\underline{52.65} \\
\cellcolor{blue!18}\textbf{SPECTRA} & $4.88\times$ & 37.74 & 26.00 & 26.05 & 74.67 & 90.23 & 44.61 & 63.77 & 61.06 & \cellcolor{blue!18}\textbf{53.02} \\
TurboQuant & $5.33\times$ & 35.92 & 25.60 & 26.03 & 75.33 & 91.61 & 46.11 & 58.75 & 59.03 & 52.30 \\
RotateKV & $5.33\times$ & 34.48 & 25.19 & 26.37 & 75.33 & 89.06 & 45.07 & 62.53 & 59.69 & 52.21 \\
SPECTRA & $5.50\times$ & 37.67 & 25.29 & 26.38 & 74.67 & 89.45 & 44.34 & 62.76 & 58.88 & 52.43 \\
\midrule
\cellcolor{blue!18}\textbf{SPECTRA} & $6.31\times$ & 37.70 & 25.93 & 25.82 & 75.33 & 89.70 & 44.51 & 62.79 & 61.78 & \cellcolor{blue!18}\textbf{52.94} \\
\cellcolor{blue!7}\underline{SPECTRA} & $6.81\times$ & 35.95 & 25.62 & 26.19 & 74.00 & 89.79 & 44.59 & 62.91 & 61.24 & \cellcolor{blue!7}\underline{52.54} \\
KIVI & $7.11\times$ & 36.09 & 24.94 & 25.67 & 74.00 & 89.84 & 46.17 & 61.60 & 56.36 & 51.83 \\
OTT & $7.11\times$ & 35.64 & 24.68 & 26.16 & 74.00 & 90.01 & 46.28 & 62.63 & 58.47 & 52.23 \\
\midrule
\cellcolor{blue!18}\textbf{SPECTRA} & $7.73\times$ & 36.29 & 25.95 & 25.97 & 72.00 & 89.41 & 45.56 & 58.38 & 55.81 & \cellcolor{blue!18}\textbf{51.17} \\
\cellcolor{blue!7}\underline{RotateKV} & $8.00\times$ & 30.24 & 24.79 & 26.25 & 70.00 & 87.31 & 42.88 & 59.87 & 53.64 & \cellcolor{blue!7}\underline{49.37} \\
TurboQuant & $8.00\times$ & 34.54 & 22.92 & 25.53 & 68.67 & 87.94 & 44.40 & 54.62 & 50.23 & 48.61 \\
\midrule
\cellcolor{blue!18}\textbf{SPECTRA} & $8.93\times$ & 32.59 & 25.84 & 25.99 & 74.00 & 88.39 & 45.56 & 56.85 & 54.05 & \cellcolor{blue!18}\textbf{50.41} \\
\midrule
\cellcolor{blue!18}\textbf{SPECTRA} & $11.28\times$ & 33.61 & 24.96 & 25.28 & 70.00 & 87.87 & 44.42 & 52.61 & 52.03 & \cellcolor{blue!18}\textbf{48.85} \\
\bottomrule\end{tabular}}\end{table}

%% file: sections/tab_qwen_longbench.tex
\begin{table}[H]\centering\small\setlength{\tabcolsep}{3.5pt}
\caption{\textbf{LongBench per task --- Qwen2.5-7B-Instruct.} Dense (fp16) and \spectra{} (best
configuration per compression tier; no baselines were run for this model). Higher is better; Avg is
the mean over the eight tasks. The \colorbox{blue!18}{\textbf{best}} and
\colorbox{blue!7}{\underline{second}} \spectra{} Avg are highlighted.}
\label{tab:lb-qwen2.5-7b}\resizebox{\textwidth}{!}{%
\begin{tabular}{l c cccccccc c}\toprule
Method & Comp. & Qasper & QMSum & MNews & TREC & TQA & SAMSum & LCC & RB-P & Avg \\\midrule
base (fp16) & $1.00\times$ & 41.61 & 23.69 & 23.91 & 68.00 & 89.40 & 45.65 & 61.97 & 64.75 & \textbf{52.37} \\\midrule
SPECTRA-LR$+$B & $2.48\times$ & 40.29 & 23.74 & 23.47 & 70.00 & 88.70 & 44.68 & 60.81 & 62.39 & 51.76 \\
SPECTRA-LR$+$B & $2.80\times$ & 40.40 & 23.72 & 23.69 & 70.00 & 88.70 & 45.02 & 60.32 & 63.43 & 51.91 \\
\cellcolor{blue!18}\textbf{SPECTRA-WF$+$B} & $3.48\times$ & 41.36 & 24.05 & 23.70 & 69.33 & 88.74 & 45.61 & 62.35 & 65.08 & \cellcolor{blue!18}\textbf{52.53} \\
SPECTRA-LR$+$B & $3.96\times$ & 40.42 & 23.40 & 23.51 & 68.00 & 89.38 & 45.48 & 55.68 & 57.17 & 50.38 \\
SPECTRA-WF$+$B & $4.44\times$ & 40.47 & 23.94 & 23.68 & 66.67 & 87.35 & 44.83 & 60.73 & 63.04 & 51.34 \\
\cellcolor{blue!7}\underline{SPECTRA-WF} & $4.92\times$ & 40.81 & 23.82 & 23.29 & 71.33 & 87.84 & 45.00 & 64.76 & 61.27 & \cellcolor{blue!7}\underline{52.27} \\
SPECTRA-LR & $6.10\times$ & 38.19 & 23.14 & 23.76 & 65.33 & 88.17 & 44.82 & 62.40 & 59.92 & 50.72 \\
\bottomrule\end{tabular}}\end{table}

%% file: sections/fig_llama_niah.tex
\begin{figure}[H]
\centering
\includegraphics[width=\linewidth]{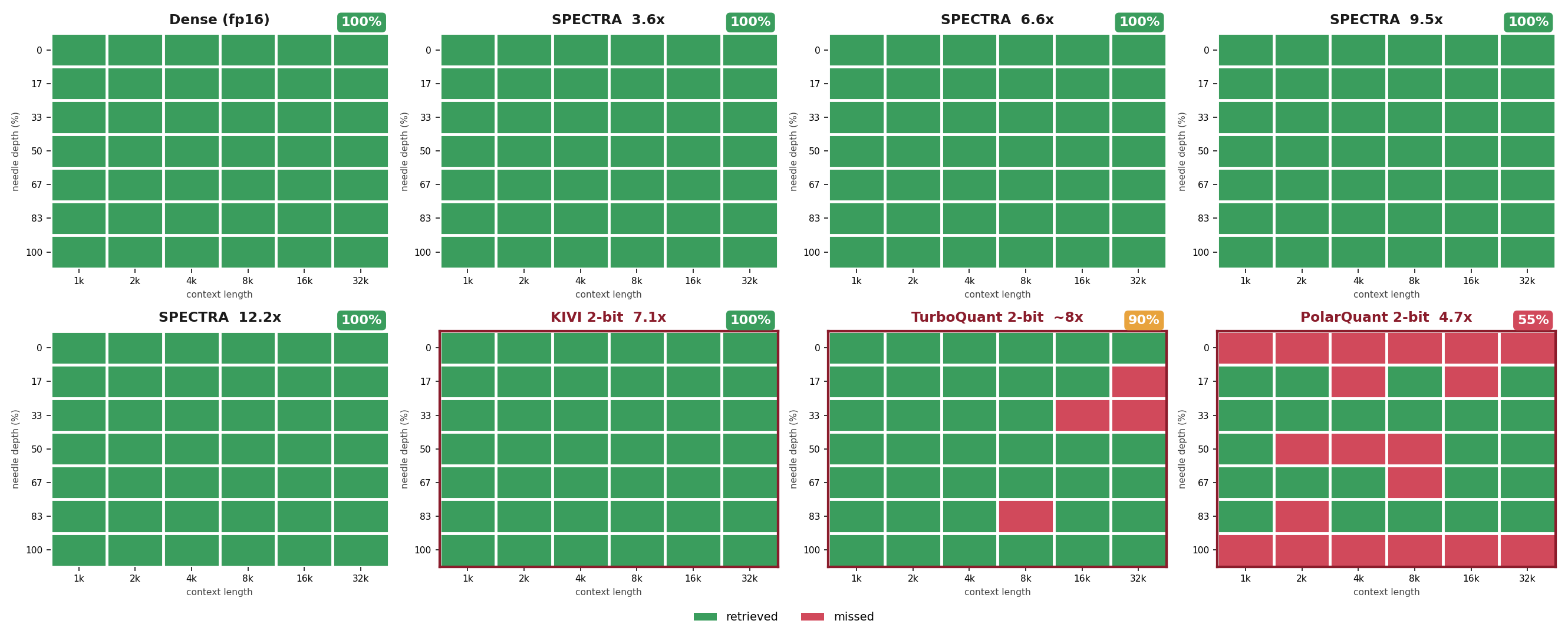}
\caption{\textbf{Single-needle NIAH --- Llama-3.1-8B-Instruct.} Retrieval accuracy across context length (x-axis) and needle depth (y-axis); green marks a retrieved needle, red a missed one, and the badge reports overall accuracy. SPECTRA matches the dense fp16 model up to high compression, staying near-perfect where 2-bit baselines (KIVI, TurboQuant, PolarQuant) begin to fail.}
\label{fig:niah-single-llama}
\end{figure}

\begin{figure}[H]
\centering
\includegraphics[width=\linewidth]{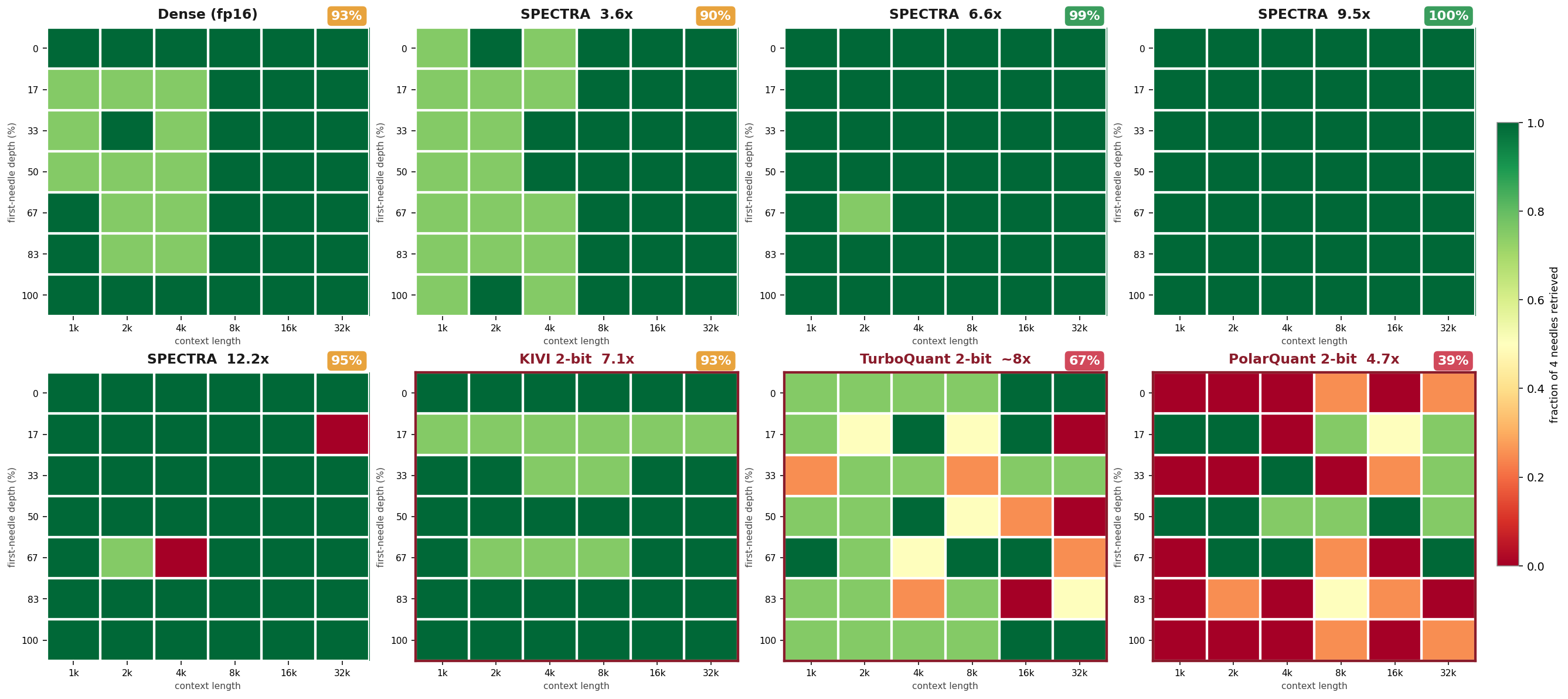}
\caption{\textbf{Multi-needle NIAH --- Llama-3.1-8B-Instruct.} Fraction of $k$ needles retrieved as a function of context length (x-axis) and needle depth (y-axis); color runs from red (none retrieved) to green (all retrieved), and the badge reports the mean fraction. SPECTRA preserves multi-needle retrieval across the depth--length grid while 2-bit baselines degrade sharply, especially at long contexts.}
\label{fig:niah-multi-llama}
\end{figure}

%% file: sections/fig_mistral_niah.tex
\begin{figure}[H]
\centering
\includegraphics[width=\linewidth]{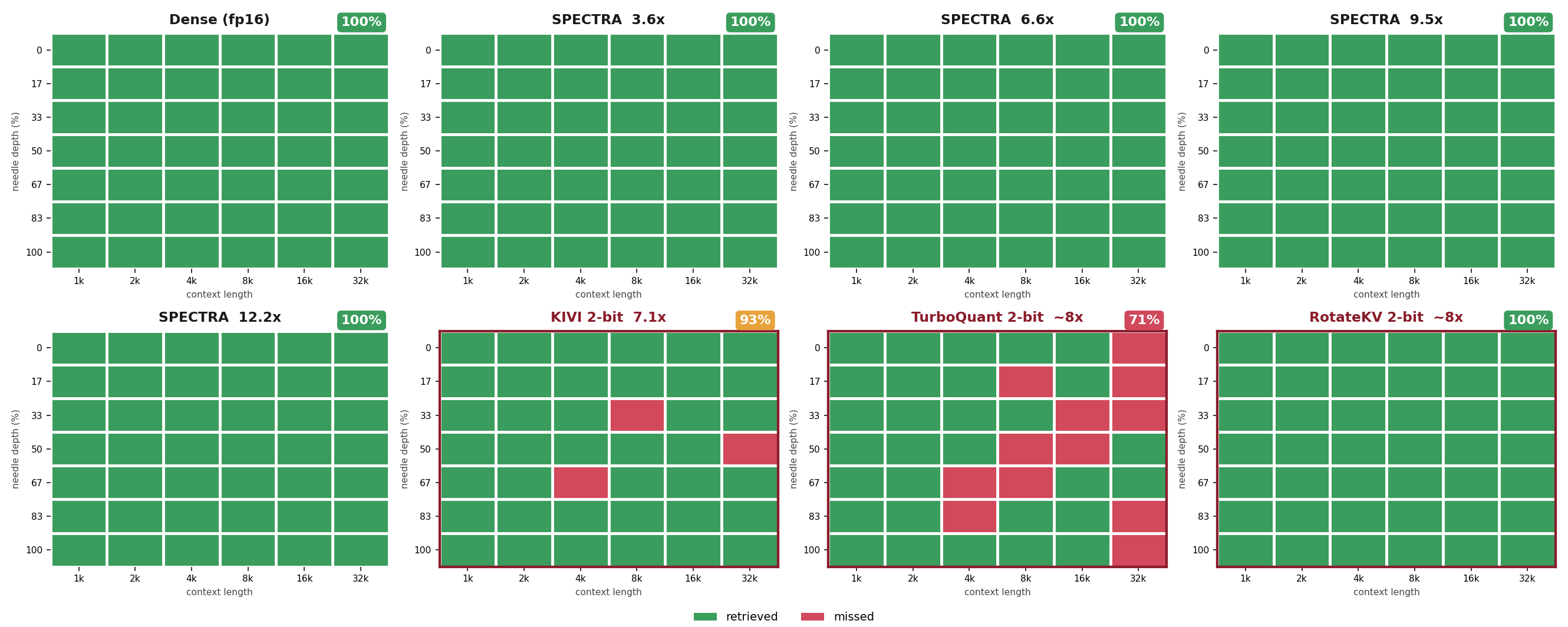}
\caption{\textbf{Single-needle NIAH --- Mistral-7B-Instruct-v0.3.} Retrieval accuracy across context length (x-axis) and needle depth (y-axis); green marks a retrieved needle, red a missed one, and the badge reports overall accuracy. SPECTRA matches the dense fp16 model up to high compression, staying near-perfect where 2-bit baselines (KIVI, TurboQuant, PolarQuant) begin to fail.}
\label{fig:niah-single-mistral}
\end{figure}

\begin{figure}[H]
\centering
\includegraphics[width=\linewidth]{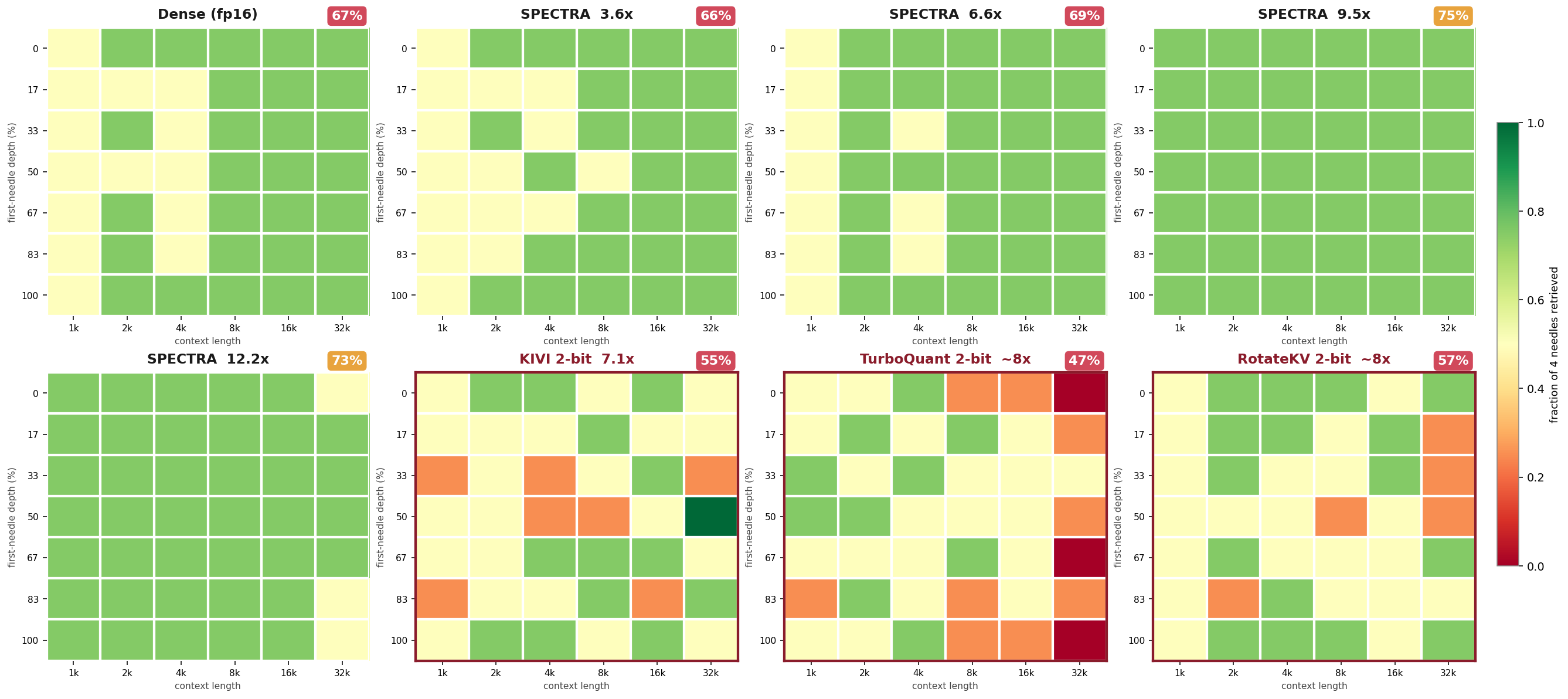}
\caption{\textbf{Multi-needle NIAH --- Mistral-7B-Instruct-v0.3.} Fraction of $k$ needles retrieved as a function of context length (x-axis) and needle depth (y-axis); color runs from red (none retrieved) to green (all retrieved), and the badge reports the mean fraction. SPECTRA preserves multi-needle retrieval across the depth--length grid while 2-bit baselines degrade sharply, especially at long contexts.}
\label{fig:niah-multi-mistral}
\end{figure}